\documentclass[11pt]{article}
	
	\newcommand{\blind}{0}
	
    \makeatletter
    \renewcommand\section{\@startsection {section}{1}{\z@}%
                                       {-3.5ex \@plus -1ex \@minus -.2ex}%
                                       {2.3ex \@plus.2ex}%
                                       {\normalfont\fontfamily{phv}\fontsize{16}{19}\bfseries}}
    \renewcommand\subsection{\@startsection{subsection}{2}{\z@}%
                                         {-3.25ex\@plus -1ex \@minus -.2ex}%
                                         {1.5ex \@plus .2ex}%
                                         {\normalfont\fontfamily{phv}\fontsize{14}{17}\bfseries}}
    \renewcommand\subsubsection{\@startsection{subsubsection}{3}{\z@}%
                                        {-3.25ex\@plus -1ex \@minus -.2ex}%
                                         {1.5ex \@plus .2ex}%
                                         {\normalfont\normalsize\fontfamily{phv}\fontsize{14}{17}\selectfont}}
    \makeatother
	\usepackage{amsmath}
    \DeclareMathOperator*{\argmax}{arg\,max}
    
    \usepackage{graphicx}
    \usepackage{algorithm}
    \usepackage{algpseudocode}
    \usepackage{braket}
    \usepackage{amssymb, amsfonts}
    \usepackage{amsthm}
    \usepackage{booktabs}
    \usepackage{adjustbox}
    \usepackage{makecell}
    
    \usepackage{adjustbox}
    
    \usepackage{siunitx}
    \usepackage{booktabs}
    \usepackage{makecell}
    \usepackage{adjustbox}
    \usepackage{float}

    \newtheorem{lemma}{Lemma}

    \newtheorem{theorem}{Theorem}
    \newtheorem{proposition}{Proposition}
    
	\usepackage{graphicx}
    \usepackage{subcaption} 
	\usepackage{enumerate}
	\usepackage{xcolor}
	\usepackage{natbib} %comment out if you do not have the package
    \newtheorem*{theoremone}{Theorem 1}
    \newtheorem*{theoremtwo}{Theorem 2}
	\usepackage{url} % not crucial - just used below for the URL
    \usepackage{algorithm}
    \usepackage{algpseudocode}
    
    \makeatletter
    \newenvironment{breakablealgorithm}[1]
      {%
       \par\addvspace{\intextsep}
       \noindent
       \refstepcounter{algorithm}%
       \hrule height 0.8pt
       \vskip 2pt
       \noindent\textbf{\ALG@name~\thealgorithm} #1\par
       \vskip 2pt
       \hrule height 0.4pt
       \vskip 4pt
      }
      {%
       \vskip 4pt
       \hrule height 0.8pt
       \par\addvspace{\intextsep}
      }
    \makeatother

\usepackage[textsize=tiny]{todonotes}
\usepackage{xspace}
\usepackage{tikz}
\usetikzlibrary{arrows.meta,positioning,fit,calc,shadows.blur}

\begin{document}
		
			%%%%%%%%%%%%%%%%%%%%%%%%%%%%%%%%%%%%%%%%%%%%%%%%%%%%%%%%%%%%%%%%%%%%%%%%%%%%%%
		\def\spacingset#1{\renewcommand{\baselinestretch}%
			{#1}\small\normalsize} \spacingset{1}
		%%%%%%%%%%%%%%%%%%%%%%%%%%%%%%%%%%%%%%%%%%%%%%%%%%%%%%%%%%%%%%%%%%%%%%%%%%%%%%
		
		\if0\blind
		{
			\title{\bf \emph{Quantum Safe-Set Bayesian Optimization for Quality Improvement in Fuselage Assembly}}
			\author{Jiayu Liu$^a$,  Chong Liu$^b$, Trevor Rhone$^c$, and Yinan Wang$^a$ \\
            Emails: \texttt{\{liuj35, rhonet, wangy88\}@rpi.edu, cliu24@albany.edu}       \\
			$^a$ Department of Industrial and Systems Engineering, Rensselaer Polytechnic Institute \\
             $^b$ Department of Computer Science, University at Albany, State University of New York \\
             $^c$ Department of Physics, Applied Physics \& Astronomy, Rensselaer Polytechnic Institute } 
			\date{}
			\maketitle
		} \fi
		
		\if1\blind
		{

            \title{\bf \emph{IISE Transactions} \LaTeX \ Template}
			\author{Author information is purposely removed for double-blind review}
			
\bigskip
			\bigskip
			\bigskip
			\begin{center}
				{\LARGE\bf \emph{IISE Transactions} \LaTeX \ Template}
			\end{center}
			\medskip
		} \fi
		\bigskip
		
	\begin{abstract}
Aircraft fuselage assembly relies on shape control, where force-controlled actuators adjust one fuselage section to reduce its dimensional gap to the adjoining section. In practice, this optimization is simultaneously safety-constrained, expensive, and noisy: (1) candidate actuator forces must improve dimensional accuracy without violating structural safety limits, (2) each evaluation requires a costly simulation or physical measurement, and (3) sensing error, calibration drift, and environmental variation corrupt each measurement, so the observed response deviates from its true mean and repeated measurements are needed to estimate it reliably. Existing Bayesian Optimization (BO) approaches for shape control do not jointly ensure safety and sample efficiency under such noise. The central bottleneck is mean-response estimation, for which the classical Monte Carlo estimation requires \(\mathcal{O}(1/\epsilon^2)\) samples to achieve estimation accuracy \(\epsilon\). This cost is prohibitive when evaluations are scarce. This motivates Quantum Monte Carlo (QMC) estimation, which reaches the same accuracy with only \(\mathcal{O}(1/\epsilon)\) samples (a quadratic reduction). In addition, the current practice of QMC estimation does not take safety constraints into consideration, which is critical when optimizing physical systems. Therefore, we propose a Quantum Safe-Set Bayesian Optimization (Q-Safe BO) framework for precise, safe, and sample-efficient shape control during assembly. The proposed framework integrates Safe-Set Bayesian Optimization with QMC estimation. The Safe-Set BO component selects actuator-force configurations that are likely to improve dimensional accuracy while satisfying the safety requirement. QMC estimates the mean response with fewer samples than classical MC methods for a target accuracy. A customized Upper Confidence Bound (UCB) acquisition strategy further balances safe exploration and objective improvement. Theoretically, the proposed framework reduces the number of samples required for response estimation while maintaining reliable constrained optimization performance. In the case study of the fuselage shape control, the experimental results demonstrate that Q-Safe BO achieves a lower dimensional gap between adjacent fuselage sections than classical methods under the same optimization budget while maintaining structural safety. Furthermore, experiments conducted on a real IBM quantum computer show that our proposed method remains competitive with its simulator-based counterpart and outperforms the classical baseline, providing an initial validation of the framework's deployability on current quantum hardware.
	\end{abstract}
			
	\noindent%

	%\newpage
	\spacingset{1.5} % DON'T change the spacing!

\section{Introduction} \label{sec:intro}
Over the past century, the materials utilized in the aerospace industry have evolved from traditional wood and metal to advanced composite materials. Composite materials, such as fiberglass-reinforced plastics, offer substantial benefits over their predecessors, including higher specific modulus, exceptional corrosion resistance, and ease of manufacturing. In modern aerospace manufacturing, large foundational components are first produced independently and then assembled to construct an entire aircraft. For instance, multiple fuselage sections are manufactured separately in the factory and subsequently joined together. A critical challenge during this assembly process is the presence of a dimensional gap between the sections being joined. Such a gap can undermine the structural stability of the aircraft. Therefore, precise shape control is crucial to maintaining the integrity and reliability of aerospace structures.

\begin{figure}[H]
    \centering
    \includegraphics[width=0.7\linewidth]{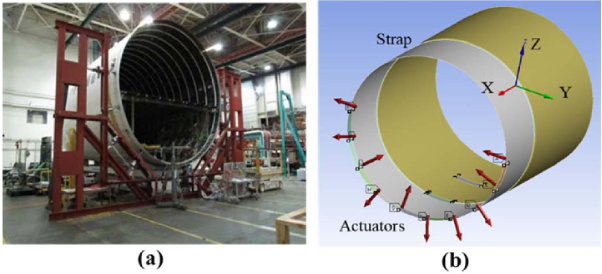}
    \caption{Schemes of (a) force-controlled actuators system \citep{WEN2018272} and (b) fixtures in fuselage assembly \citep{yuesurrogate}.}
    \label{fig:fuselage}
\end{figure} 

Shape control during fuselage assembly typically involves the installation of force-controlled actuators along the edges of fuselage sections to correct shape distortions by applying either pulling or pushing forces. The optimal number and positioning of these actuators are determined through experimental design procedures. Fig. \ref{fig:fuselage} provides a schematic illustration of the shape control process, highlighting how one fuselage section remains fixed while the adjacent section is precisely aligned using multiple force-controlled actuators. These actuators are adjusted to minimize the gaps between sections, with their forces set in the radial direction within the fuselage coordinate system either inward (push) or outward (pull). In addition to reducing dimensional gap, the applied actuator forces must also satisfy the structural safety requirement. Since the fuselage sections are made of composite materials, excessive or improper force combinations may cause material failure. Therefore, the force combinations are constrained by the Tsai–Wu failure criterion \citep{tsai1971general}, which is used in this study to ensure that only structurally safe loading conditions are considered during shape control.

Identifying effective actuator placements and force configurations in fuselage assembly is challenging because the problem involves structural safety constraints, costly evaluations, and noisy measurements. First, not every actuator-force configuration is physically feasible. A force setting may reduce the dimensional gap between fuselage sections while violating the structural safety requirement, such as the Tsai–Wu failure criterion. Therefore, fuselage shape control should be formulated as a constrained optimization problem that minimizes the dimensional gap while ensuring structural safety. Second, each evaluation of an actuator configuration can be expensive. Real-world experiments require physical measurements, and simulation-based evaluations using Finite Element Analysis (FEA) can also be time-consuming. Third, the measured response of a configuration is uncertain due to sensing errors, calibration drift, and environmental variations. Under such noise, estimating the mean dimensional response of a given actuator configuration to a prescribed precision may require many repeated observations. These challenges motivate the need for an optimization framework that can reduce costly evaluations, estimate noisy responses efficiently, and maintain structural safety throughout the sequential search process.

Existing studies have addressed individual aspects of fuselage shape-control optimization, but they have not jointly resolved the three practical challenges of safety constraints, expensive evaluations, and measurement noise. Surrogate models have been developed to approximate FEA-based dimensional-gap evaluations and thereby reduce the computational cost of assessing actuator configurations \citep{yuesurrogate}. Reinforcement Learning (RL) has also been used to learn policies that generalize across different initial fuselage shapes \citep{Lutz19112024}. Although RL reduces the need to solve a new optimization problem for each initial condition, it requires extensive offline training and a large number of interaction samples. More importantly, neither the surrogate-model-based framework nor the RL-based framework explicitly enforces structural safety constraints, such as the Tsai–Wu failure criterion. To address the lack of explicit structural-feasibility consideration in earlier methods, Bayesian Optimization (BO) has been investigated for selecting actuator configurations under structural constraints \citep{albahar2022stress, albahar}. In particular, \citet{albahar2022stress} incorporated the Tsai--Wu failure criterion using Constrained Expected Improvement (CEI). CEI combines the expected improvement of a candidate configuration with its predicted probability of feasibility. It therefore favors actuator configurations that are both promising and likely to satisfy the structural constraint. However, CEI does not construct a high-confidence safe set or restrict every evaluation to such a set. Consequently, CEI accounts for structural feasibility when searching for an optimum, but it does not explicitly guarantee the safety of every intermediate actuator configuration evaluated during the sequential optimization process.

Beyond fuselage shape control, BO has been widely used in manufacturing process optimization when evaluations are expensive and only a limited number of experiments can be conducted. For example, constrained BO has been used to optimize turning and grinding parameters subject to process-quality and operating constraints \citep{MAIER201981,maier2020self}. More recently, \citet{li2026bayesian} proposed Bayesian Optimization with Active Constraint Learning (BO-ACL) for advanced manufacturing process design. BO-ACL combines objective optimization with active constraint learning by selecting informative samples based on representativeness, uncertainty, and diversity to improve the estimation of an initially unknown feasible region.

However, these BO-based manufacturing optimization methods are primarily designed to identify high-performing feasible solutions or to estimate feasibility boundaries, rather than to guarantee safe exploration throughout the entire sequential optimization process. Safe exploration imposes a stronger requirement: every configuration evaluated during optimization must satisfy the prescribed safety requirement with high confidence. For example, BO-ACL may evaluate configurations near an uncertain feasibility boundary because such observations are informative for constraint learning. Although this strategy is useful for learning the feasible region, it may be unacceptable in fuselage shape control, where evaluating an unsafe actuator configuration could violate structural limits or damage the physical system. Therefore, the distinction between constraint-aware optimization and safe exploration is particularly important for safety-critical fuselage assembly. In addition to this safety requirement, measurement noise introduces another practical challenge. Although BO frameworks commonly model noisy observations, noisy manufacturing measurements may still require repeated evaluations of the same actuator configuration to estimate its true mean response with sufficient accuracy. This repeated-sampling requirement increases the effective evaluation cost and motivates the need for efficient mean-response estimation under noisy observations.

Taken together, existing surrogate-model-based, reinforcement-learning-based, and BO-based methods address parts of this problem, but they do not jointly provide sample-efficient actuator selection, efficient estimation of noisy mean responses, and high-confidence safe exploration. To address these gaps, we propose a Quantum Safe-Set Bayesian Optimization (Q-Safe BO) framework for the fuselage shape-control problem. The proposed framework integrates surrogate modeling, Quantum Monte Carlo (QMC) estimation, and safe-set exploration. Specifically, surrogate models are used to approximate costly objective and safety evaluations associated with actuator-force configurations. The safe-set component constructs a set of actuator configurations that are predicted to satisfy the Tsai–Wu failure criterion with high confidence. The acquisition function is then optimized only within this safe set, allowing the method to search for configurations with a smaller dimensional gap while restricting sequential evaluations to safe regions. The QMC estimation addresses the cost of estimating noisy responses. For a fixed actuator configuration, classical Monte Carlo (MC) estimation requires $\mathcal{O}(1/\epsilon^2)$ response samples to estimate the mean response to accuracy $\epsilon$, whereas QMC estimation requires $\mathcal{O}(1/\epsilon)$ samples under the same accuracy requirement \citep{montanaro2015quantum}. Therefore, under a fixed total query budget, QMC estimation can either improve the accuracy of response estimation at each actuator configuration or allow more configurations to be evaluated during the optimization process. This provides more informative observations for updating the surrogate models and selecting subsequent safe configurations. Motivated by recent Quantum Bayesian Optimization (QBO) results, which establish improved cumulative regret under quantum oracle access \citep{dai2023quantum}, we examine the solution quality under the same total query budget. Specifically, we examine whether Q-Safe BO achieves lower cumulative regret and identifies actuator configurations with a smaller dimensional gap than classical methods while maintaining the Tsai–Wu failure criterion throughout the sequential optimization process.

We summarize our contributions as follows:

\begin{itemize}
\item We propose a Q-Safe BO framework for fuselage shape control that addresses the sample-efficiency requirement in costly manufacturing optimization. The framework integrates surrogate modeling with QMC estimation, where surrogate models reduce the need for repeated FEA-based evaluations and QMC estimation lowers the sampling cost required to estimate noisy mean responses.

\item We develop a safe sequential decision-making strategy that addresses the safety requirement in fuselage assembly. By constructing a nominal safe set from the Tsai--Wu failure criterion, the proposed method restricts sequential evaluations to actuator-force configurations that satisfy structural feasibility with high confidence.

\item We establish theoretical guarantees for the proposed framework: cumulative regret bounds under bounded-noise and bounded-variance observation models (Theorems~\ref{thm:objective_confidence_regret} and \ref{thm:modified_bounded_variance_regret_paper}), which characterize the dependence of performance on the noise magnitude $\sigma$ and the actuator dimension $d$, and a safety guarantee that all queried configurations satisfy the Tsai--Wu constraint with high confidence (Proposition~\ref{prop:safety}).
\end{itemize}

The rest of this paper is organized as follows. First, Sec. \ref{sec:related} reviews the literature on quantum computing and fuselage assembly. Next, Sec. \ref{sec:QBOFA} and Sec. \ref{sec:simulated} present the quantum computing preliminaries, the proposed method, and the simulation study. Then, Sec. \ref{sec:experiment} describes the experimental settings and reports the results. Finally, Sec. \ref{sec:conclusion} concludes the paper and summarizes the main findings. For clarity, we distinguish among BO stages, samples in classical BO, and queries/iterations in quantum oracle. A BO stage emphasizes determining the control variable $\mathbf{F}$ and its response. 
A classical sample refers to one noisy response observation obtained from a physical measurement, an FEA-based simulator, or a surrogate model. 
A quantum query/iteration refers to one access to the quantum reward oracle, i.e., one call of \(\mathcal{O}_{\mathbf{F}}\) or \(\mathcal{O}_{\mathbf{F}}^{\dagger}\), for a selected force vector \(\mathbf{F}\). 
In the QBO setting, one quantum query/iteration is treated as the quantum counterpart of one classical reward sample or one classical arm pull. 
Therefore, the cumulative regret of QBO is defined with respect to the total number of quantum oracle queries/iterations, not the number of BO stages.

\section{Related Work} \label{sec:related}
Designing and optimizing manufacturing processes often involves iterative exploration of a large solution space while balancing accuracy, cost, and robustness. 
In fuselage assembly, achieving precise shape control requires both accurate simulation of structural responses and minimizing the number of measurements in a noisy environment. These challenges motivate researchers to seek advanced algorithms. Recent studies indicate that quantum algorithms can achieve accuracy and sample efficiency comparable to, or even exceeding, those of classical methods in certain manufacturing tasks. In this section, we review related works in two parts: (i) methods for quality control and unconstrained/constrained optimization in fuselage assembly processes, and (ii) quantum algorithms that offer potential advantages for computationally intensive simulation and optimization tasks in industry and related fields. 

\subsection{Quality Control in Aircraft Assembly Process}
In manufacturing, design and process optimization are commonly formulated as the problem of finding an optimal solution over a predefined design space. This process can be viewed as an interaction between a solver and an environment: the solver strategically generates candidate designs, while the environment evaluates each candidate and returns quantitative feedback. Therefore, the effectiveness of manufacturing optimization depends on both the fidelity of the environment model and the sample efficiency of the optimization strategy.

In recent years, significant research efforts have been devoted to composite fuselage assembly. Early studies mainly focused on constructing high-fidelity virtual environments to simulate the physical assembly process. For example, \cite{WEN201955} proposed an FEA method to virtually simulate the assembly of two composite fuselage sections and analyze the stress induced by actuator-applied forces. To account for uncertainty during shape control, \cite{yuesurrogate} developed a surrogate model to improve the accuracy and efficiency of simulation-based evaluation. To further reduce the computational burden of repeated FEA or surrogate-model evaluations, \cite{Zhong} introduced a novel FEA framework specifically designed for fuselage shape control. Together, these studies established the virtual environments needed to evaluate actuator-force configurations in fuselage assembly.

Based on these virtual environments, optimization methods have been developed to determine actuator forces that reduce the dimensional gap between fuselage sections. \cite{duOptimalPlacementActuators2019} constructed a linear model to describe the relationship between actuator forces and dimensional gap, and proposed an optimization-based method for gap minimization. Extending this direction, \cite{duNewSparseLearningModel2022} introduced a sparse learning model with an $l_{\infty}$ loss function to further improve dimensional control. In addition to deterministic optimization approaches, reinforcement learning and BO have also been explored for gap reduction \citep{Lutz19112024, albahar}. Moreover, \cite{Mou02112023} introduced a sparse-sensor-based adaptive control method to improve shape control performance. These studies show that optimization and learning-based methods can improve fuselage assembly quality by sequentially searching for better actuator-force configurations.

However, fuselage assembly is not simply an unconstrained gap-minimization problem. In practical manufacturing systems, a candidate design may improve the target quality metric but still be unacceptable if it violates physical, material, or operational constraints. This issue is particularly important in composite fuselage assembly, where actuator forces used to reduce the dimensional gap may also increase local stress, residual deformation, or material failure risk. Therefore, actuator-force planning should be formulated as a constrained optimization problem, in which the objective is to minimize the dimensional gap while satisfying the structural safety requirement.

Recent studies have begun to incorporate such constraint-related considerations into fuselage assembly. For example, stress-aware actuator placement methods account for residual stress during actuator selection and force adjustment, demonstrating that dimensional-gap reduction must be balanced with structural integrity \citep{albahar2022stress}. More broadly, in advanced manufacturing, BO with active constraint learning has been proposed to jointly optimize process parameters and learn feasibility constraints \citep{li2026bayesian}. These studies highlight the importance of learning both the objective response and the feasible design region in manufacturing.

Nevertheless, fuselage assembly imposes a stricter requirement on constrained optimization under the noisy condition: the optimizer must be sample-efficient. Each actuator-force evaluation can be expensive because it may require FEA simulation, surrogate-model evaluation, or physical measurement. Meanwhile, measurement noise reduces the reliability of the observed dimensional gap, and structural safety constraints restrict the feasible search region. As a result, the optimizer must learn both the objective response and the safe design region from a limited number of evaluations. This motivates us to develop a sample-efficient constrained optimization framework for noisy composite fuselage assembly.

\subsection{Quantum Algorithms in Advanced Manufacturing}
Advanced simulation and data-driven modeling techniques, such as the
finite element method (FEM) and deep learning-based models, have been
investigated and developed as the ``environment''
\citep{wang2024smartfixture, wang2024multimodal}, and optimization
methods, such as mathematical programming, have been proposed as the
``solver'' \citep{duOptimalPlacementActuators2019, duNewSparseLearningModel2022}. From the “environment” perspective, compared with its classical counterpart, quantum computing has enhanced the simulation capabilities of scientific computing, particularly in developing general linear algebra routines (a core subroutine in FEM) \citep{clader2013preconditioned, montanaro2016quantum}. Linear systems are ubiquitous in engineering, naturally arising in applications such as PDE solving. Classical algorithms typically require $\mathcal{O}(\chi\sqrt{\kappa})$ time to solve a linear system, where $\kappa$ is the condition number and $\chi$ is the matrix dimension. The Harrow–Hassidim–Lloyd (HHL) algorithm, however, can achieve $\mathcal{O}(\log(\chi)\kappa^{2}/\epsilon)$ time complexity \citep{harrow2009quantum}. Such quantum linear solvers have been proposed for accelerating large-scale FEM simulations in structural mechanics and materials modeling \citep{montanaro2016quantum, clader2013preconditioned}, potentially reducing computation times in design and analysis phases of advanced manufacturing.

From the “solver” perspective, quantum computing has been explored in operations research for diverse optimization tasks. Quantum Approximate Optimization Algorithm (QAOA) and variational quantum eigensolvers have been applied to combinatorial scheduling problems, such as the Job-Shop Scheduling Problem (JSSP) \citep{farhi2014quantum, kurowski2023application, venturelli2016job}. In addition to scheduling, quantum algorithms have been investigated for quality control and defect detection in manufacturing. Quantum machine learning methods, such as Quantum Support Vector Machines (QSVM), have been applied to anomaly detection and defect classification in industrial manufacturing \citep{guijo2022quantum}, in some cases outperforming classical baselines on unbalanced datasets. Quantum-assisted anomaly detection has also been proposed for predictive maintenance of gas turbines in power plants \citep{sakhnenko2022hybrid}. These studies illustrate that quantum computing has been shown to be effective in manufacturing. In the context of fuselage assembly, these quantum capabilities can potentially address two major challenges. First, quantum linear solvers offer the potential to accelerate FEM-based simulations, which dominate the cost of evaluating dimensional gap, thereby enabling more rapid exploration of the design space. Second, quantum optimization algorithms such as QAOA may improve the efficiency of searching for optimal actuator forces. Together, these capabilities could greatly benefit actuator placement, enhance robustness to uncertainty, and support real-time quality control during aircraft assembly.

\section{Quantum Safe-Set Bayesian Optimization for Fuselage Assembly}\label{sec:QBOFA}
\subsection{Problem Definition and Formulation}
\label{sec:problem_formulation}
In aircraft assembly, a dimensional gap inevitably occurs when two fuselage sections are joined because of a shape mismatch between the sections. Composite materials used in aircraft construction are more flexible than traditional materials such as metal and wood, which makes fuselage sections easier to deform during shape control. Therefore, force-controlled actuators are commonly installed along the fuselage edges to push or pull one section toward the desired target shape. The goal is to determine actuator-force combinations that reduce the dimensional gap while satisfying the structural safety requirement.

In this work, we formulate fuselage shape control as a constrained optimization problem following \citet{albahar2022stress}. The decision variable is the actuator-force vector
\begin{equation}
    \mathbf{F}
    =
    (F_1,F_2,\ldots,F_m)^{\top}
    \in \mathcal{A}
    =
    \left\{
    \mathbf{F}\in\mathbb{R}^m:
    F_{\min}\leq F_j\leq F_{\max},
    \ j=1,\ldots,m
    \right\},
    \label{eq:force_bound}
\end{equation}
where \(m\) is the number of actuators and \(\mathcal{A}\) is the admissible force domain.
Let \(N\) denote the number of measurement points along the fuselage cross-section. The initial shape is represented by
\begin{equation}
    \mathbf{P}_0
    =
    \left(
    \mathbf{p}_0^1,
    \mathbf{p}_0^2,
    \ldots,
    \mathbf{p}_0^N
    \right),
\end{equation}
where each point is $\mathbf{p}_0^i=(x_0^i,y_0^i,z_0^i), i=1,\ldots,N$. Similarly, the target shape is represented by
\begin{equation}
    \mathbf{P}_t
    =
    \left(
    \mathbf{p}_t^1,
    \mathbf{p}_t^2,
    \ldots,
    \mathbf{p}_t^N
    \right),
\end{equation}
where $\mathbf{p}_t^i=(x_t^i,y_t^i,z_t^i)$. When the actuator-force vector \(\mathbf{F}\) is applied, the fuselage section deforms. 
Let
\begin{equation}
    \mathbf{U}(\mathbf{F})
    =
    \left(
    \mathbf{u}^1(\mathbf{F}),
    \mathbf{u}^2(\mathbf{F}),
    \ldots,
    \mathbf{u}^N(\mathbf{F})
    \right)
\end{equation}
denote the displacement induced by \(\mathbf{F}\), where \(\mathbf{u}^i(\mathbf{F})\) is the displacement of the \(i\)-th measurement point. In real manufacturing environments, the measured response is affected by measurement noise caused by sensor error, calibration uncertainty, temperature-induced deformation, and other sources of uncertainty. We define the noise at each point

\begin{equation}
    \boldsymbol{\xi}
    =
    \left(
    \boldsymbol{\xi}^1,
    \boldsymbol{\xi}^2,
    \ldots,
    \boldsymbol{\xi}^N
    \right).
\end{equation}
The measured final shape under \(\mathbf{F}\) is then written as
\begin{equation}
    \mathbf{P}_f(\mathbf{F})
    =
    \mathbf{P}_0
    +
    \mathbf{U}(\mathbf{F})
    +
    \boldsymbol{\xi}.
    \label{eq:final_shape_noise}
\end{equation}
For the \(i\)-th measurement point, the initial dimensional gap and final measured dimensional gap with respect to the target shape are defined as
\begin{equation}
\begin{aligned}
    \boldsymbol{\delta}_0^i
    &=
    \mathbf{p}_t^i-\mathbf{p}_0^i, \\
    \boldsymbol{\delta}_f^i
    &=
    \mathbf{p}_t^i-\mathbf{p}_f^i.
\end{aligned}
\label{eq:deviation_constrained}
\end{equation}
Using Eq. \eqref{eq:final_shape_noise}, the final measured dimensional gap can be written equivalently as
\begin{equation}
    \boldsymbol{\delta}_f^i(\mathbf{F})
    =
    \boldsymbol{\delta}_0^i
    -
    \mathbf{u}^i(\mathbf{F})
    -
    \boldsymbol{\xi}^i.
\end{equation}
If the noise is defined directly in the dimensional gap domain, the same model can be written as
\begin{equation}
    \boldsymbol{\delta}_f^i(\mathbf{F})
    =
    \boldsymbol{\delta}_0^i
    -
    \mathbf{u}^i(\mathbf{F})
    +
    \tilde{\boldsymbol{\xi}}^i,
\end{equation}
where \(\tilde{\boldsymbol{\xi}}^i=-\boldsymbol{\xi}^i\). In the numerical experiments, the measurement noise is modeled as additive Gaussian noise with 0 mean and a known standard deviation.

The dimensional gap after shape control is evaluated by the Mean Absolute Error (MAE):
\begin{equation}
    \mathcal{L}(\mathbf{F})
    =
    \frac{1}{N}
    \sum_{i=1}^{N}
    \left\|
    \boldsymbol{\delta}_f^i(\mathbf{F})
    \right\|,
    \label{eq:mae_constrained}
\end{equation}
where $N$ is the number of measurement points. Because the response is noisy, the same actuator-force vector \(\mathbf{F}\) may produce different observed values of \(\mathcal{L}(\mathbf{F})\). Therefore, the objective of BO is to improve the expected performance on fuselage shape control:
\begin{equation}
    f(\mathbf{F})
    =
    \mathbb{E}
    \left[
    -\mathcal{L}(\mathbf{F})
    \right].
    \label{eq:expected_objective_constrained}
\end{equation}
The negative sign converts the dimensional-gap minimization problem into a maximization problem (to meet the needs of BO). A larger value of \(f(\mathbf{F})\) corresponds to a smaller dimensional gap.

In addition to reducing the dimensional gap, the actuator-force vector must meet the structural safety requirement of the composite material. Excessive or improper actuator forces can introduce stress states that lead to material failure. We use the Tsai--Wu failure criterion \citep{tsai1971general} to define this safety requirement. Let \(\mathrm{FI}(\mathbf{F})\)
denote the Tsai--Wu failure index associated with the stress field induced by \(\mathbf{F}\). A force vector is structurally safe if \(\mathrm{FI}(\mathbf{F})\leq 1\).
Equivalently, we define the safety function
\begin{equation}
    g(\mathbf{F})
    =
    1-\mathrm{FI}(\mathbf{F}).
    \label{eq:safety_margin}
\end{equation}
Thus, \(\mathbf{F}\) is safe if \(g(\mathbf{F})\geq 0.\)
Combining the dimensional-gap objective, actuator-force bounds, and Tsai--Wu failure criterion, the fuselage shape-control problem is formulated as
\begin{equation}
\begin{aligned}
    \mathbf{F}^{*}
    &=
    \arg\max_{\mathbf{F}\in\mathcal{A}}
    \quad
    f(\mathbf{F})
    =
    \mathbb{E}
    \left[
    -\mathcal{L}(\mathbf{F})
    \right] \\
    \mathrm{s.t.}
    \quad
    &F_{\min}\leq F_j\leq F_{\max},
    \qquad j=1,\ldots,m, \\
    &g(\mathbf{F})=1-\mathrm{FI}(\mathbf{F})\geq 0.
\end{aligned}
\label{eq:constrained_fuselage_problem}
\end{equation}

This formulation differs from an unconstrained gap-minimization problem because a force vector with a small dimensional gap may still be infeasible if it violates the Tsai--Wu failure criterion. Therefore, the optimizer must simultaneously search for force vectors that improve shape alignment and remain within the structurally safe region.

\subsection{Quantum Safe-Set Bayesian Optimization}
\subsubsection{Overview}
The proposed Q-Safe BO framework consists of three modules: the quantum oracle, the quantum mean estimation subroutine, and the Safe-Set BO loop. First, for a selected actuator-force vector, the FEA-based model or surrogate model provides the deterministic, or mean, structural response. The prescribed noise model then defines a response distribution around this mean response. The quantum oracle formalizes idealized coherent access to this noise-induced response distribution. Second, the quantum mean estimation subroutine estimates the expected scalar response associated with the selected force vector. Third, the Safe-Set BO module uses the estimated response and the safety observation to update the Gaussian Process models and select the next safe and informative actuator-force vector. For completeness, Appendix~\ref{appdix:clsBO} provides a brief overview of unconstrained BO. While unconstrained BO does not consider physical constraints in optimizing force configurations, fuselage assembly has a specific safety requirement to ensure structural feasibility, which is demonstrated in Eq. \eqref{eq:constrained_fuselage_problem}. 

\begin{figure}[!ht]
  \centering
    \includegraphics[width=1.0\textwidth]{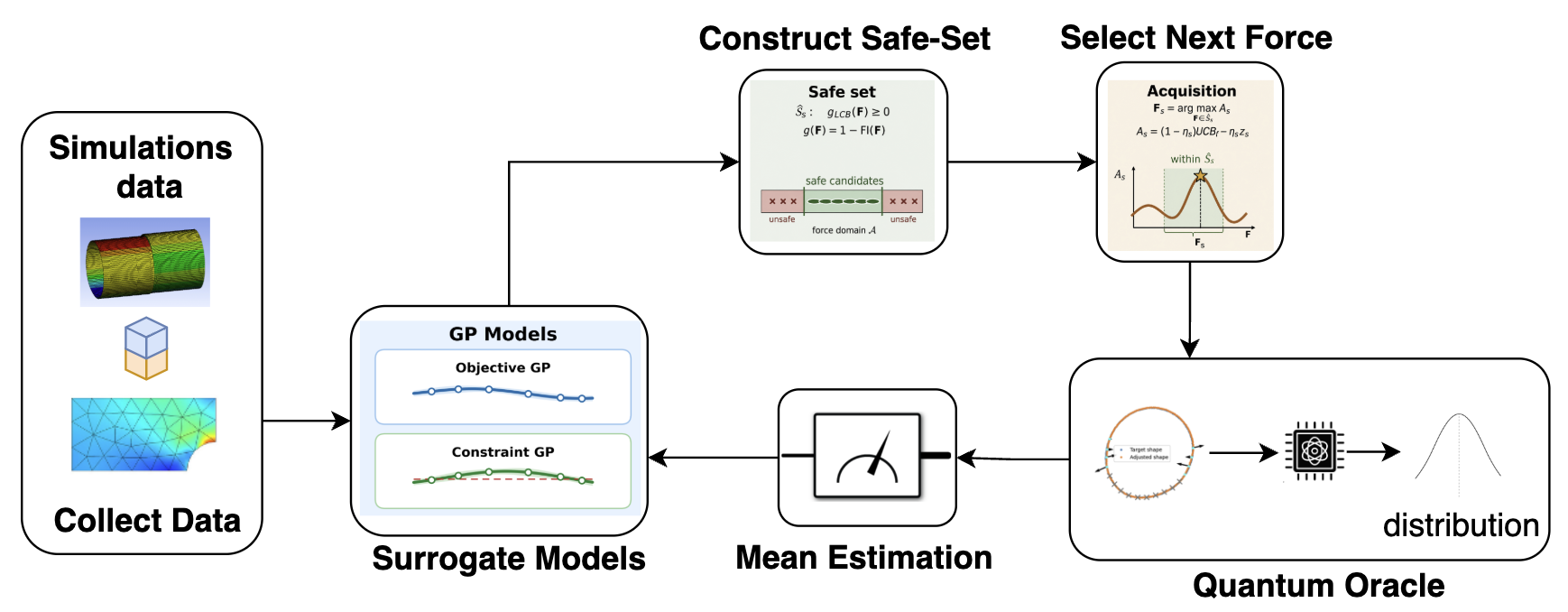}
  \caption{Pipeline of the proposed Q-Safe BO framework. FEA simulation data are first used to train surrogate models for the structural response and safety function. The acquisition function selects the next actuator-force vector in the nominal safe set. For the selected force vector, QMC estimation estimates the expected scalar response, which is then used to update the BO surrogate model.}
  \label{fig:pipe}
\end{figure}

The overall pipeline is shown in Fig.~\ref{fig:pipe}. FEA simulation data are first used to train surrogate models for the structural response and safety function. At each BO stage, the acquisition function selects a candidate force vector from the nominal safe set. The QMC estimation then estimates the expected response of this candidate. Finally, the new observation is used to update the BO surrogate models, and the process is repeated.

\subsubsection{Quantum Oracle and Access Model}
\label{sec:QO}

The quantum oracle provides the formal access model between the FEA or surrogate-based response model and the quantum mean estimation subroutine. For a given actuator-force vector \(\mathbf{F}\), the FEA-based model or surrogate model provides the deterministic, or mean, dimensional gap. The dimensional gap distribution is then defined by adding the prescribed measurement-noise model to this mean gap. Thus, the stochasticity considered in this work is not generated by the FEA solver itself. Instead, it is introduced through the noise model around the FEA or surrogate predicted mean gap.

Following the standard QBO formulation \citep{dai2023quantum}, the action of the force-conditioned quantum oracle is defined as
\begin{equation}
    \mathcal{O}_{\mathbf{F}}:
    |\mathbf{F}\rangle |0\rangle_{\Omega}|0\rangle_{Y}
    \rightarrow
    |\mathbf{F}\rangle
    \sum_{\omega\in\Omega_{\mathbf{F}}}
    \sqrt{P_{\mathbf{F}}(\omega)}
    |\omega\rangle_{\Omega}
    |y_{\mathbf{F}}(\omega)\rangle_{Y},
    \label{eq:qoracle_force_conditioned}
\end{equation}
where \(|\mathbf{F}\rangle\) is the force vector, \(|\omega\rangle_{\Omega}\) represents a possible noisy outcome, \( \sqrt{P_{\mathbf{F}}(\omega)}\) is the probability amplitude of the outcome, and \(|y_{\mathbf{F}}(\omega)\rangle_{Y}\) stores the corresponding scalar response. For continuous noise distributions, \(\Omega_{\mathbf{F}}\) is interpreted as a finite-precision discretization of possible noisy outcomes. When required by the QMC estimation, the scalar response is shifted and scaled into the appropriate bounded range for quantum encoding.

This access model should be interpreted carefully. Quantum mean estimation requires access to both \(\mathcal{O}_{\mathbf{F}}\) and its inverse \(\mathcal{O}_{\mathbf{F}}^{\dagger}\). A standard classical FEA solver does not provide such access, because it evaluates one force vector and returns one response per run. Therefore, the quantum speedup discussed in this paper is an oracle-query complexity advantage under the standard coherent-oracle model. It is not a claim that a classical FEA code can be directly executed in quantum superposition.

Under this convention, one quantum oracle query means one invocation of \(\mathcal{O}_{\mathbf{F}}\) or \(\mathcal{O}_{\mathbf{F}}^{\dagger}\) for the selected force vector. It is not one BO stage and is not one direct call to a classical FEA solver. Instead, it is treated as the quantum counterpart of one classical noisy reward sample from the response distribution. This query accounting allows Q-Safe BO regret to be compared with BO regret while keeping the idealized quantum access model separate from the FEA-based response model.

\subsubsection{Quantum Mean Estimation}
\label{sec:ME} 
For a candidate force vector \(\mathbf{F}\), the objective value is defined as
\(f(\mathbf{F})=\mathbb{E}[y_{\mathbf{F}}]\). 
Because this expectation is not directly available, Monte Carlo estimation is required to approximate \(f(\mathbf{F})\) from repeated evaluations. In Q-Safe BO, this estimation step is performed using Quantum Monte Carlo instead of classical Monte Carlo.

\begin{lemma}
\label{Lemma:QMC}
(Quantum Monte Carlo method \citep{montanaro2015quantum}. Adapted from QBO \citep{dai2023quantum})
Let \(y_x:\Omega_x\to\mathbb{R}\) denote a random variable, \(\Omega_x\) is equipped with probability measure \(P_x\), and the quantum unitary oracle \(\mathcal{O}_x\) encodes \(P_x\) and \(y_x\).

\noindent 
\textbf{Bounded Noise}: If the noisy output observation satisfies \(y_x\in[0,1]\), then there exists a constant \(C_1>1\) and a QMC estimation algorithm \(QMC(\mathcal{O}_x,\epsilon,\delta)\) which returns an estimate \(\hat y_x\) of \(\mathbb{E}[y_x]\) such that \(\mathbb{P}\left(|\hat y_x-\mathbb{E}[y_x]|>\epsilon\right)\leq \delta,\) using at most \(\frac{C_1}{\epsilon}\log\left(\frac{1}{\delta}\right)\) queries to \(\mathcal{O}_x\) and its inverse.
    
\noindent 
\noindent 
\textbf{Noise with Bounded Variance}: 
If the variance of \(y_x\) satisfies \(\mathrm{Var}(y_x)\leq \sigma^2\), then for 
\(\epsilon<4\sigma\), there exists a constant \(C_2>1\) and a QMC estimation algorithm 
\(QMC(\mathcal{O}_x,\epsilon,\delta)\) which returns an estimate \(\hat y_x\) of 
\(\mathbb{E}[y_x]\) such that
\(
\mathbb{P}\left(|\hat y_x-\mathbb{E}[y_x]|>\epsilon\right)\leq \delta .
\)
The algorithm uses at most \(N_{\mathrm{QMC}}^{\mathrm{var}}\) queries to 
\(\mathcal{O}_x\) and its inverse, where
\(
N_{\mathrm{QMC}}^{\mathrm{var}}
=
\frac{C_2\sigma}{\epsilon}
\log_2^{3/2}\left(\frac{8\sigma}{\epsilon}\right)
\log_2\left(\log_2\frac{8\sigma}{\epsilon}\right)
\log\left(\frac{1}{\delta}\right).
\)
\end{lemma}

This result shows the query-complexity advantage of QMC estimation. Suppose that the variance at the response level is bounded by \(\mathrm{Var}(y)\leq \sigma_y^2\). For a classical sample mean computed from \(k\) independent noisy samples, Chebyshev's inequality gives
\(
    \mathbb{P}
    \left(
    |\hat{\mu}-\mu|\geq \epsilon
    \right)
    \leq
    \frac{\sigma_y^2}{k\epsilon^2}.
\)
Thus, to guarantee an error at most \(\epsilon\) with failure probability at most \(\delta\), the classical estimator requires
\(
    k
    =
    O\left(
    \frac{\sigma_y^2}{\delta\epsilon^2}
    \right)
\)
samples. In contrast, QMC estimation requires only \(O(1/\epsilon)\) oracle queries up to logarithmic factors and problem-dependent constants. Therefore, QMC estimation reduces the dependence on the target accuracy from quadratic to nearly linear in \(1/\epsilon\).

As an illustrative comparison, suppose that the mean and standard deviation of the response are $\mu=0.5$ and \(\sigma=0.25\), the target precision is \(\epsilon=0.01\), and the confidence level is \(1-\delta=0.95\). Chebyshev's inequality indicates the classical estimation requires $k \geq \frac{\sigma^2}{\delta\epsilon^2} = \frac{0.25^2}{0.05\times 0.01^2} = 12,500$ samples. Under the bounded noise assumption in Lemma \ref{Lemma:QMC}, the number of quantum oracle queries scales approximately as $O\left(\frac{C_1}{\epsilon} \log\left( \frac{1}{\delta}\right)\right)$, which is nearly linear in \(1/\epsilon\) up to logarithmic factors. This improved scaling is the main reason Q-Safe BO can reduce the number of queries needed for accurate mean estimation.

\subsubsection{Safe BO Loop with QMC Estimation}
\label{sec:safeset_qbo}

In this section, we extend the QBO and Weighted Bayesian Optimization (WBO) frameworks \citep{dai2023quantum,deng2022weighted} to the constrained fuselage assembly problem. In particular, a force combination that improves dimensional accuracy may still be infeasible if it violates the material failure constraint \citep{tsai1971general}. Therefore, the optimizer must account for both the objective response and the safety constraint during sequential decision-making.

To address this requirement, the proposed Q-Safe BO maintains two surrogate models. The first model is a Gaussian Process for the safety function \(g(\mathbf{F})\), which represents the structural feasibility condition. The second model is a weighted Gaussian Process for the objective function \(f(\mathbf{F})\), which represents the dimensional-gap response. The objective surrogate model guides the optimizer toward actuator-force combinations that reduce the dimensional gap, while the safety model restricts the search away from force combinations that are likely to violate the Tsai--Wu failure criterion.

The constraint observation at a selected force vector is computed from the Tsai--Wu failure index:
\begin{equation}
    g(\mathbf{F_s})
    =
    1-\mathrm{FI}(\mathbf{F}_s).
\end{equation}
We use a Gaussian Process, denoted by $\hat{g}$, as the surrogate model for the safety function:
\begin{equation}
\label{fun:surrogatesafe}
    \hat{g}(\mathbf{F})
    \sim
    GP\left(
    \mu_{g,s-1}(\mathbf{F}),
    \sigma_{g,s-1}^{2}(\mathbf{F})
    \right).
\end{equation}
The nominal safe set, $\widehat{S}_{s}$, at stage \(s\) is constructed using the lower confidence bound of the surrogate model in Eq. \eqref{fun:surrogatesafe}:
\begin{equation}
    \mathrm{LCB}_{\hat{g},s}(\mathbf{F})
    =
    \mu_{g,s-1}(\mathbf{F})
    -
    \beta_s^{(\mathrm{con})}
    \sigma_{g,s-1}(\mathbf{F}),
    \label{eq:lcb_constraint}
\end{equation}
and
\begin{equation}
    \widehat{S}_s
    =
    \left\{
    \mathbf{F}\in\mathcal{A}:
    \mathrm{LCB}_{\hat{g},s}(\mathbf{F})\geq 0
    \right\}.
    \label{eq:safe_set}
\end{equation}

The lower confidence bound requires
$\mu_{g,s-1}(\mathbf{F})
-
\beta_s^{(\mathrm{con})}\sigma_{g,s-1}(\mathbf{F})
\geq 0$. This means that the force vector remains feasible even under a conservative realization of the uncertain safety function. Uncertainty ($\sigma_{g,s-1}$) therefore provides a safety margin against surrogate-model prediction errors. Equivalently, a candidate must satisfy
$\mu_{g,s-1}(\mathbf{F})
\geq\beta_s^{(\mathrm{con})}\sigma_{g,s-1}(\mathbf{F})$, so force vectors with larger predictive uncertainty require a larger positive predicted safety margin before they can be included in the nominal safe set.

% Hence, every force vector in \(\widehat{S}_s\) satisfies the Tsai--Wu failure criterion with high confidence.}

After constructing the nominal safe set, we then build the objective surrogate:

\begin{equation}\label{eq:objective_surrogate}
\tilde{f}(\mathbf{F})\sim GP(\tilde{\mu}_{f,s-1},\tilde{\sigma}_{f,s-1}),
\end{equation}
where $\tilde{\mu}_{f,s-1}$ and $\tilde{\sigma}_{f,s-1}$ are the weighted GP posterior mean and standard deviation of the objective surrogate.
The next force vector is selected by maximizing an acquisition function over $\widehat{S}_s$. The acquisition function is defined as
\begin{equation}
A_s(\mathbf{F})
=
(1-\eta_s)\mathrm{UCB}_{f,s}(\mathbf{F})
-
\eta_s z_s(\mathbf{F}),
\label{eq:safeset_acquisition}
\end{equation}
where
\begin{equation}
\eta_s=\frac{1}{s+1}.
\label{eq:eta_schedule}
\end{equation}
The acquisition function has two components: the first term is called Upper Confidence Bound (UCB), which is defined as 
\begin{equation}
\mathrm{UCB}_{f,s}(\mathbf{F})
=
\tilde{\mu}_{f,s-1}(\mathbf{F})
+
\beta_s^{(f,c)}
\tilde{\sigma}_{f,s-1}(\mathbf{F}).
\label{eq:ucb_objective}
\end{equation}
To encourage safety exploration, we also define a safety-boundary exploration term
\begin{equation}
z_s(\mathbf{F})
=
\left|
\frac{
\mu_{g,s-1}(\mathbf{F})
}{
\sigma_{g,s-1}(\mathbf{F})
}
\right|.
\label{eq:z_term}
\end{equation}
This quantity measures the standardized distance from the estimated safety boundary. 
At each stage, the algorithm selects
\begin{equation}
\mathbf{F}_s
=
\arg\max_{\mathbf{F}\in \widehat{S}_s}
(1-\eta_s)\mathrm{UCB}_{f,s}(\mathbf{F})
-
\eta_s z_s(\mathbf{F}).
\label{eq:safeset_selection}
\end{equation}
The first term promotes force configurations with promising objective values or high objective uncertainty, while the second term encourages safe exploration near the estimated constraint boundary. A smaller value of $z_s(\mathbf{F})$ indicates that the force vector is either closer to the safety boundary or located in a region where the constraint model remains uncertain. Since $\eta_s=\frac{1}{s+1}$ decreases with the stage number, the algorithm places more emphasis on safe exploration in the early stages and gradually shifts toward objective optimization in later stages.

Both surrogate models in the proposed Q-Safe BO are equipped with confidence events. Let $\mathcal{E}_g$ denote the event that $|g(\mathbf{F})-\mu_{g,s-1}(\mathbf{F})|\le\beta_s^{(\mathrm{con})} \sigma_{g,s-1}(\mathbf{F})$ for all $\mathbf{F}\in\mathcal{A}$ and all $s$, and let $\mathcal{E}_f$ denote the analogous event for the weighted objective model, $|f(\mathbf{F})-\tilde{\mu}_{f,s-1}(\mathbf{F})|\le\beta_s^{(f,c)} \tilde{\sigma}_{f,s-1}(\mathbf{F})$ for all $\mathbf{F}\in\mathcal{A}$ and all $s$. Following \citet{dai2023quantum}, each event holds with probability at least $1-\delta_g$ and $1-\delta$, respectively, for appropriately chosen confidence parameters $\beta_s^{(\mathrm{con})}$ and $\beta_s^{(f,c)}$ (specified in Appendix~\ref{Appdix:B}). The following result formalizes the safety guarantee.
\begin{proposition}[Safe exploration]
\label{prop:safety}
With probability at least $1-\delta_g$, i.e., on the event $\mathcal{E}_g$, every configuration selected by Algorithm~1 satisfies $g(\mathbf{F}_s)\ge 0$ simultaneously for all stages $s$.
\end{proposition}
\begin{proof}
By construction of the nominal safe set, $\mathbf{F}_s\in\widehat{S}_s$ implies $\mathrm{LCB}_{\hat g,s}(\mathbf{F}_s)=\mu_{g,s-1}(\mathbf{F}_s)-\beta^{(\mathrm{con})}_s\sigma_{g,s-1}(\mathbf{F}_s)\ge 0$. On $\mathcal{E}_g$, $g(\mathbf{F})\ge \mathrm{LCB}_{\hat g,s}(\mathbf{F})$ for all $\mathbf{F}\in\mathcal{A}$ and all $s$; hence $g(\mathbf{F}_s)\ge 0$ for every $s$. Since $\mathbb{P}(\mathcal{E}_g)\ge 1-\delta_g$, the claim follows.
\end{proof}

At stage $s$, suppose that the selected force vector is \(\mathbf{F}_s\), QMC estimation is used to estimate the expected objective response \(f(\mathbf{F}_s)\). The QMC estimation precision is chosen according to the current weighted posterior uncertainty of the objective model ($\tilde{\sigma}_{f,s-1}(\mathbf{F}_{s})$):
\begin{equation}
\epsilon_s
=\min\big(
c
\frac{
\tilde{\sigma}_{f,s-1}(\mathbf{F}_s)
}{
\sqrt{\lambda}
},\epsilon_{\max}\big)
\label{eq:qmc_precision_safe}
\end{equation}
where \(c\in(0,1]\) is the precision-scaling constant, $\epsilon_{\max}$ is the maximum allowed estimation error, and $\lambda$ is the regularization parameter. When the posterior uncertainty at \(\mathbf{F}_s\) is large, a coarser estimate is sufficient. When the posterior uncertainty is small, a more accurate QMC estimation is used. Thus, more accurate QMC estimates receive larger weights in the weighted GP update. After \(s\) stages, the objective dataset is
\begin{equation}
\mathcal{D}^{f}_{s}
=
{(\mathbf{F}_i,\hat{y}_i,w_i)}_{i=1}^{s},
\end{equation}
where \(\hat{y}_i\) is the QMC estimation at \(\mathbf{F}_i\), and \(w_i\) is the corresponding observation weight defined as $w_i = \frac{1}{\epsilon_i^2}$. Following the weighted GP formulation, we define
\begin{equation}
\label{eq:Vfs}
V_{f,s}
=
\lambda I
+
\sum_{i=1}^{s}
w_i\phi(\mathbf{F}_i)\phi(\mathbf{F}_i)^{\top},
\end{equation}
where \(\lambda\) is the regularization parameter and \(\phi(\cdot)\) is the Random Fourier Feature (RFF) mapping \citep{rahimi2007random}. Note that this $\lambda$ is the same as in Eq.~\eqref{eq:qmc_precision_safe}.

The full procedure is summarized in Algorithm~\ref{alg:safeset_qbo}. We also conducted the theoretical analysis showing that the proposed 
Q-Safe BO algorithm enjoys sublinear regret. Specifically, let $\widetilde{R}_T^{(c)}=\sum_{s=1}^{m^{(c)}}N_s\bigl(f(\mathbf{F}_s^\dagger)-f(\mathbf{F}_s)\bigr)$ denote the nominal-set cumulative regret, where \(N_s\) is the number of oracle queries used by the QMC estimator at stage \(s\), \(\mathbf{F}_s\) is the selected safe candidate, \(\mathbf{F}_s^\dagger\) is the best candidate in the nominal safe set at stage \(s\), $m^{(c)}$ is the number of stages completed within the query budget $T$. \(\mathbf{F}_s^\dagger\) is used here to calculate the cumulative regret, as in practice, the optimal solution is not always available in a constrained optimization problem. We provide an upper bound for nominal-set cumulative regret in Theorem~\ref{thm:objective_confidence_regret}, with the full proof and regularity conditions included in Appendix~\ref{Appdix:B}. Throughout the theoretical analysis, $d$ denotes the dimension of the search domain. For example, in the fuselage shape-control problem, $d$ equals the number of active actuators.

\begin{theorem}[Bounded noise regret of Q-Safe BO]
\label{thm:objective_confidence_regret}
Assume that the objective confidence event \(\mathcal E_f\) holds. For a fixed 
precision-scaling constant \(c\) and a fixed confidence level \(\delta\), the 
cumulative regret of Q-Safe BO satisfies
\[
\widetilde{R}_{T}^{(c)}
=
\mathcal{O}\left(
\frac{
(\log T)^{\frac{3(d+1)}{2}}
\log\bigl((\log T)^{d+1}\bigr)
}
{
c\log(1+c^{-2})
}
\right)
+
\mathcal{O}\left(
\frac{1}{c}
\bigl(\log((\log T)^{d+1})\bigr)^2
\right).
\]
\end{theorem}
Theorem~\ref{thm:objective_confidence_regret} characterizes the regret of 
Q-Safe BO under the prescribed QMC precision with bounded noise. However, in fuselage 
assembly, the oracle response has noise with bounded variance. Therefore, the number of oracle queries \(N_s\) required to achieve the target precision depends on the magnitude of the observation noise variance. To make this dependence explicit, we 
state a bounded variance regret result below.

\begin{theorem}[Bounded variance regret of Q-Safe BO]
\label{thm:modified_bounded_variance_regret_paper}
Suppose the maximum estimation error satisfies \(\epsilon_{\max}<4\sigma\), so that the bounded-variance query bound of Lemma~\ref{Lemma:QMC} applies at every stage. Then, with probability at least \(1-\delta\), for the squared-exponential (SE) kernel, the cumulative regret satisfies
\[
\begin{aligned}
\widetilde R_T^{(c)}
&=
\mathcal O\!\left(
\frac{
\sigma\,(\log T)^{3(d+1)/2}
\log\!\bigl((\log T)^{d+1}\bigr)
\log_2^{3/2}(\sigma T)
\log_2\!\bigl(\log_2(\sigma T)\bigr)
}{
c\,\log(1+c^{-2})
}
\right)
\\
&\quad
+
\mathcal O\!\left(
\frac{
\sigma
\log_2^{3/2}(\sigma T)
\log_2\!\bigl(\log_2(\sigma T)\bigr)
}{
c
}
\bigl(\log\!\bigl((\log T)^{d+1}\bigr)\bigr)^2
\right).
\end{aligned}
\]
For the linear kernel, the regret satisfies
\[
\begin{aligned}
\widetilde R_T^{(c)}
&=
\mathcal O\!\left(
\frac{
\sigma\,d^{3/2}(\log T)^{3/2}
\log(d\log T)
\log_2^{3/2}(\sigma T)
\log_2\!\bigl(\log_2(\sigma T)\bigr)
}{
c\,\log(1+c^{-2})
}
\right)
\\
&\quad
+
\mathcal O\!\left(
\frac{
\sigma
\log_2^{3/2}(\sigma T)
\log_2\!\bigl(\log_2(\sigma T)\bigr)
}{
c
}
\bigl(\log(d\log T)\bigr)^2
\right).
\end{aligned}
\]
\end{theorem}

Theorem~\ref{thm:modified_bounded_variance_regret_paper} extends Theorem~\ref{thm:objective_confidence_regret} by explicitly incorporating \(\sigma\). The bound shows that the cumulative regret scales linearly with \(\sigma\), up to logarithmic factors. Thus, larger observation noise requires more QMC oracle queries to achieve the same estimation precision, which increases the accumulated regret. This provides a theoretical explanation for the experimental trend that higher noise levels lead to larger cumulative regret. The full proof of Theorem~\ref{thm:modified_bounded_variance_regret_paper} is included in Appendix \ref{Appdix:c}, under the regularity conditions of Appendix~\ref{Appdix:B}.

\begin{breakablealgorithm}{Quantum Safe-Set Bayesian Optimization}
\label{alg:safeset_qbo}
\begin{algorithmic}[1]
\Require Gaussian Processes for objective model $f$, constraint model $g$ as surrogate models, total queries budget $T$, regularization $\lambda$, confidence bounds $\beta_s^{(f,c)}$, $\beta_s^{(con)}$, the maximum allowed estimation error $\epsilon_{\max}$, and a constant $c \in (0,1]$.
\State Initialize GP posteriors with existing data. Set query counter $q \gets 0$.
\For{stages $s = 1, 2, \dots$}
    \State \textbf{Step 1: Construct the nominal safe set}
    \State Define $\widehat{S}_s = \{\mathbf{F} \in \mathcal{A} \mid \mu_{g,s-1}(\mathbf{F}) - \beta_s^{(con)}\sigma_{g,s-1}(\mathbf{F}) \ge 0\}$
    \State \textbf{Step 2: Compute Acquisition Function}
    \State Set mixing coefficient: $\eta_s$
    \State Compute safety-boundary exploration term: $z_s(\mathbf{F}) = \left| \frac{\mu_{g,s-1}(\mathbf{F})}{\sigma_{g,s-1}(\mathbf{F})} \right|$ for all $\mathbf{F} \in \widehat{S}_s$
    \State Compute acquisition: $A_s(\mathbf{F}) = (1-\eta_s)\mathrm{UCB}_{f,s}(\mathbf{F}) - \eta_s z_s(\mathbf{F})$
    \State \textbf{Step 3: Select and Evaluate}
    \State Select candidate: $\mathbf{F}_s = \arg\max_{\mathbf{F} \in \widehat{S}_s} A_s(\mathbf{F})$
    \State Set target precision: $\epsilon_s = \min\left\{c\frac{\tilde{\sigma}_{f,s-1}(\mathbf{F}_s)}{\sqrt{\lambda}},\ \epsilon_{\max}\right\}$
    \State Compute required queries $N_s$ for precision $\epsilon_s$ \Comment{e.g., $N_s = \frac{C_1}{\epsilon_s}\log\frac{2\bar{m}^{(c)}}{\delta}$}
    \If{$q + N_s > T$}
        \State \textbf{break} \Comment{remaining budget insufficient; terminate}
    \EndIf
    \State Run $QMC(\mathcal{O}, \epsilon_s, \delta)$ once to obtain estimate $\hat{y}_s$ at $\mathbf{F}_s$; set $q \gets q + N_s$
    \State Evaluate $g_s = g(\mathbf{F}_s)$
    \State \textbf{Step 4: Model Update}
    \State Update the GP posteriors for $f$ and $g$ using the new observation $(\mathbf{F}_s, \hat{y}_s, w_s, g_s)$.
\EndFor
\end{algorithmic}
\end{breakablealgorithm}

\section{Simulation Study}\label{sec:simulated}
In the simulation study, we show the performance of  Q-Safe BO using a two-dimensional synthetic constrained optimization problem. Following \citet{basudhar2012constrained}, the objective function is defined as
\begin{equation}
    f(\mathbf{x}) = x_{1}^{2} - \sin(4x_{2}^{2}),
    \qquad
    \text{s.t. } g(\mathbf{x}) = x_{2}-x_{1}^{2}\geq 0,
\end{equation}
where \(\mathbf{x}=(x_1,x_2)\) is defined over the domain \([-1,1]\times[-1,1]\).

This benchmark is useful for constrained optimization methods because the unconstrained objective contains two symmetric minimum points, but only one of them is associated with a feasible solution that meets the constraint. Therefore, an effective constrained optimization method must not only identify points with optimal objective performance but also avoid selecting candidates that violate the constraint. For numerical evaluation, we discretize the search domain into a \(25\times25\) grid, leading to 625 candidate points, where the feasible global optimum value is $f(\mathbf{x}^{*})=-0.979.$ This example provides a controlled setting to test whether the proposed Q-Safe BO method can efficiently identify the feasible optimum while respecting constraints.

We compare five methods in the experiments. The first method, C-Safe BO, is a classical GP-based Safe-Set BO algorithm with a constrained UCB acquisition. The second method, Q-Safe BO, uses the same safe-set logic but replaces classical objective evaluation with QMC estimation through Iterative Amplitude Estimation (IAE) \citep{grinko2021iterative}. The third method, Q-Safe BO (Real), is a hardware-adapted version of the Q-Safe BO for a real IBM quantum computer. The fourth method, unconstrained BO, is a standard UCB-based BO method that ignores the safety requirement \citep{srinivas2009gaussian}. Finally, BO-ACL is an active constraint learning baseline that jointly learns the objective and the feasibility boundary \citep{li2026bayesian}. Each objective evaluation returns a noisy observation $y(\mathbf{x}) = f(\mathbf{x}) + \varepsilon$, where $\varepsilon \sim \mathcal{N}(0, \sigma^2)$ with $\sigma = 0.3$, consistent with the noise model in Sec.~3.1. Constraint evaluations are noiseless.

To compare methods, we report three metrics. Let $\mathbf{x}^{*}$ denote the feasible global optimum (available in the simulation study) and $\mathbf{x}_s$ the configuration selected at stage $s$. The empirical cumulative regret under a budget of $T$ oracle queries is $R_T=\sum_{s} N_s\bigl(f(\mathbf{x}_s)-f(\mathbf{x}^{*})\bigr)$, where $N_s$ is the number of queries used to estimate the response at stage $s$ (classical samples for MC-based methods, oracle queries for QMC-based methods), consistent with the query-based accounting in Sec.~\ref{sec:intro}. The simple regret is $f(\hat{\mathbf{x}}_T)-f(\mathbf{x}^{*})$, where $\hat{\mathbf{x}}_T$ is the best safe configuration identified within the budget. The violation rate is the fraction of stages whose selected configuration violates the true constraint, i.e., $g(\mathbf{x}_s)<0$. 

Note that Theorems~\ref{thm:objective_confidence_regret} and \ref{thm:modified_bounded_variance_regret_paper} bound the nominal-set regret, in which the comparator $\mathbf{x}_s^{\dagger}=\arg\min_{\mathbf{x}\in\widehat{S}_s}f(\mathbf{x})$ (the analogue of $\mathbf{F}_s^{\dagger}$ in Sec.~3.2.4; the theorems apply to the equivalent maximization of $-f$) is the best configuration in the \emph{current} nominal safe set, whereas the empirical evaluation uses the fixed comparator $\mathbf{x}^{*}$. The two metrics are directly related through the decomposition\[f(\mathbf{x}_s)-f(\mathbf{x}^{*})=\underbrace{\bigl(f(\mathbf{x}_s)-f(\mathbf{x}_s^{\dagger})\bigr)}_{\text{nominal-set regret}} \;+\; \underbrace{\bigl(f(\mathbf{x}_s^{\dagger})-f(\mathbf{x}^{*})\bigr)}_{\text{safe-set gap}}. \] Whenever the constraint confidence event holds, the nominal safe set satisfies $\widehat{S}_s\subseteq\{\mathbf{x}\in\mathcal{A}:g(\mathbf{x})\ge 0\}$ (Proposition~\ref{prop:safety}), so $f(\mathbf{x}_s^{\dagger})\ge f(\mathbf{x}^{*})$ and the safe-set gap is nonnegative with high probability. Hence, the reported regret upper-bounds the nominal-set regret analyzed in Theorems~\ref{thm:objective_confidence_regret} and \ref{thm:modified_bounded_variance_regret_paper}, making the empirical evaluation a conservative test of the theory.

\subsection{Results}
\begin{figure}[!ht]
    \centering
    \includegraphics[width=0.7\columnwidth]{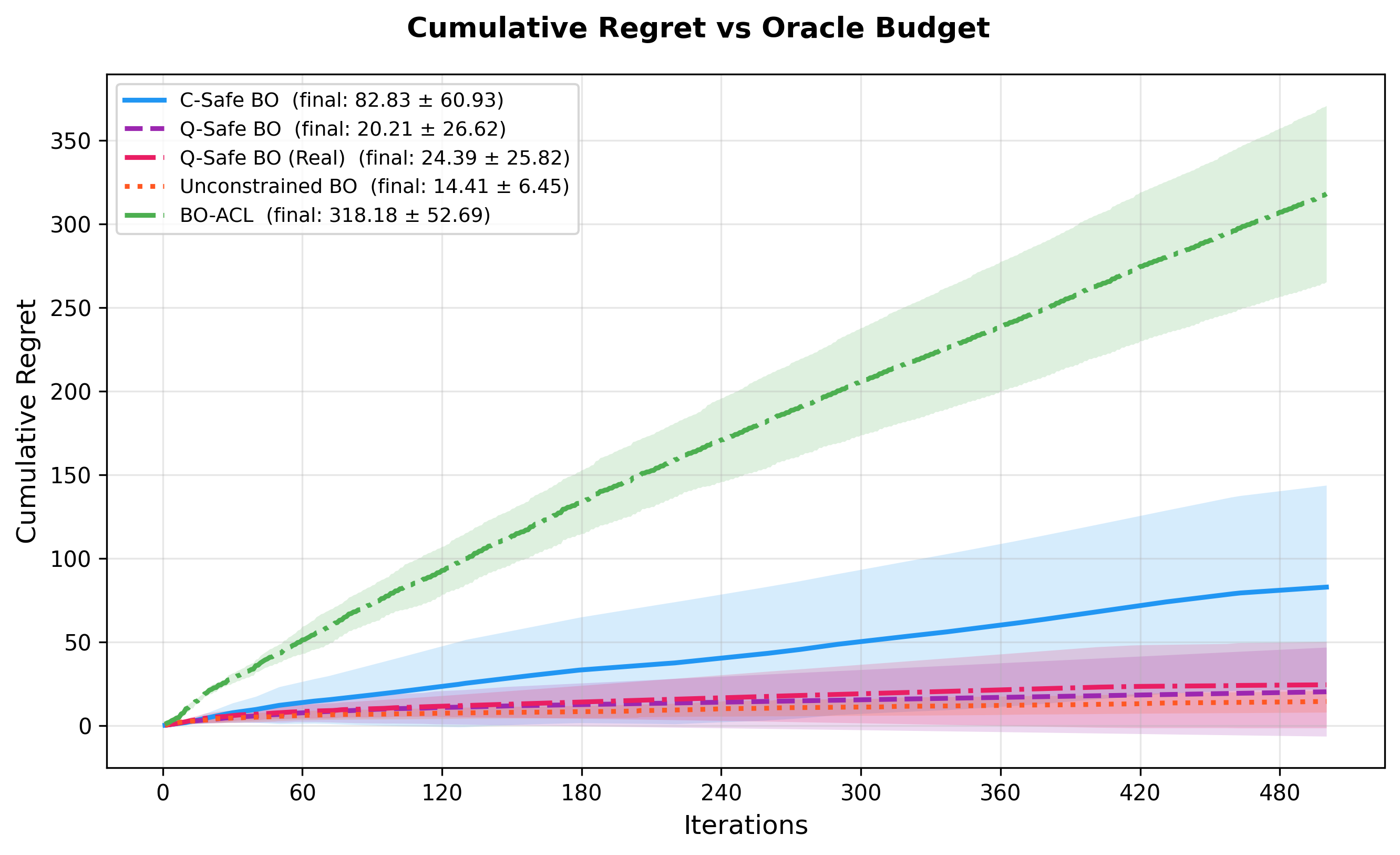}
    \caption{Mean cumulative regret curves over five independent trials. The solid or dashed curves denote the mean cumulative regret, and the shaded regions denote the confidence interval across trials.}
    \label{fig:multi_run_regret}
\end{figure}

To evaluate the robustness of different methods with respect to initialization, we conduct five independent trials with different initial point sets. We use $N_{\mathrm{init}}=5$ initial points, 500 iterations, a $25\times25$ grid, and an objective noise level of 0.3. The results are reported in Tab. ~\ref{tab:multi_seed_simulated} and the corresponding cumulative regret curves are shown in Fig. \ref{fig:multi_run_regret}.

\begin{table}[H]
\centering
\scriptsize
\setlength{\tabcolsep}{4pt}
\renewcommand{\arraystretch}{1.05}

\begin{adjustbox}{max width=\columnwidth}
\begin{tabular}{lccc}
\toprule
Method 
& \makecell{Cumul.\\Regret} 
& \makecell{
 Simple\\Regret} 
& \makecell{Violation\\Rate} \\
\midrule
C-Safe BO          
& \(82.826 \pm 60.932\)  
& \(0.057 \pm 0.097\) 
& \(\textbf{0.000} \pm 0.000\) \\

Q-Safe BO        
& \(20.205 \pm 26.616\) 
& \(0.013 \pm 0.025\) 
& \(\textbf{0.000} \pm 0.000\) \\

Q-Safe BO (Real) 
& \(24.392 \pm 25.817\) 
& \(\textbf{0.006} \pm 0.011\) 
& \(\textbf{0.000} \pm 0.000\) \\

Unconstrained BO       
& \(\textbf{14.412} \pm 6.454\) 
& \(0.090 \pm 0.151\) 
& \(0.392 \pm 0.479\) \\

BO-ACL           
& \(318.183 \pm 52.691\) 
& \(0.070 \pm 0.135\) 
& \(0.315 \pm 0.069\) \\

\bottomrule
\end{tabular}
\end{adjustbox}
\caption{Comparison over 5 independent trials for the simulation study.}
\label{tab:multi_seed_simulated}
\end{table}

Tab.~\ref{tab:multi_seed_simulated} shows that Q-Safe BO, Q-Safe BO (Real), and C-Safe BO all maintain a violation rate of \(0.000 \pm 0.000\) across five trials, indicating that their safe-set constraints prevent unsafe evaluations consistently. In contrast, unconstrained BO achieves a much higher violation rate of \(0.392 \pm 0.479\), while BO-ACL achieves \(0.315 \pm 0.069\). These results show that methods without strict safe-set enforcement may evaluate unsafe points. In terms of cumulative regret, Q-Safe BO reduces the average cumulative regret from 82.83 for C-Safe BO to 20.21. Q-Safe BO (Real) further achieves the third lowest average cumulative regret \(24.39\) at the end stage. These results indicate that Q-Safe BO achieves substantially better optimization performance than the classical constrained baselines under the same iteration budget, while maintaining feasibility at every queried configuration. Furthermore, Fig.~\ref{fig:multi_run_regret} provides the comparison of cumulative regret over the full iterations. The regret curves show that Q-Safe BO and Q-Safe BO (Real) consistently remain below C-Safe BO throughout most of the optimization process. By contrast, BO-ACL accumulates regret much more rapidly and exhibits a large gap from Q-Safe BO. This behavior suggests that, although BO-ACL attempts to learn the constraint boundary, it is less efficient in balancing objective optimization and constraint satisfaction under the same iterations.

In addition, we visualize a single representative run to better demonstrate the results. The corresponding numerical results are reported in Appendix~\ref{Appdix:singlerun}. Fig. \ref{fig:single_seed_maps} provides a spatial explanation of the methods' behavior. The first row shows the queried points over the objective landscape, where the red dashed curve represents the true feasibility boundary. C-Safe BO, Q-Safe BO, and Q-Safe BO (Real) keep their queried points inside the feasible region. Among these methods, Q-Safe BO converges more efficiently around the feasible global optimum, which explains its lower simple regret and cumulative regret. By comparison, unconstrained BO ignores the constraint model and therefore selects points in the infeasible region. This behavior explains why it violates the safety requirement at every queried point. BO-ACL actively probes the constraint boundary, but several of its queried points still fall outside the feasible region. As a result, BO-ACL learns useful constraint information but does not provide the same level of safety guarantee as the Q-Safe BO and C-Safe BO. The second and third rows of Fig. \ref{fig:single_seed_maps} compare the initial and learned feasibility probability maps. For Q-Safe BO and C-Safe BO, the learned feasible region becomes more consistent with the true constraint boundary after sequential evaluations. This demonstrates that the constraint GP progressively improves its estimate of the safe region while the objective model searches for high-performing feasible points. In contrast, unconstrained BO has no feasibility model, and BO-ACL learns the constraint boundary less conservatively, which leads to a higher violation rate.

\begin{figure}[!ht]
    \centering
    \includegraphics[width=1.0\linewidth]{singleseed.jpg}
    \caption{Query maps and learned feasibility regions. The red dashed curve denotes the true constraint boundary \(g(\mathbf{x})=x_2-x_1^2=0\). Cross markers in BO-ACL indicate points are chosen by active constraint learning, whereas circular markers indicate points are chosen by the Upper Confidence Bound acquisition function.}
    \label{fig:single_seed_maps}
\end{figure}

% \begin{table}[!htbp]

Overall, the experiment leads to two conclusions. First, strict safe-set construction is necessary for constrained optimization because it maintains feasibility throughout the search process. Second, among all methods that maintain feasibility, Q-Safe BO achieves the lowest cumulative regret under the same iterations across different initializations. These results are consistent with the design motivation of Q-Safe BO: QMC estimation improves the efficiency of noisy objective evaluation, while the safe set acquisition function restricts the search to feasible regions.

\section{Fuselage Assembly Experiments}\label{sec:experiment}
% \subsection{Experimental Setting}

\subsection{Fuselage Assembly Environment}
We evaluate the proposed method using a surrogate-model-based fuselage assembly environment, with selected actuator-force configurations further validated in a pyMAPDL-based FEA environment implemented in ANSYS Parametric Design Language \citep{ANSYS,szabo2021finite}. Because repeated FEA evaluations are computationally expensive, the surrogate model is used as the main environment during the BO process. The experimental setting follows the actuator-force configuration in \citet{Lutz19112024}. Force-controlled actuators are placed at fixed positions along the fuselage edge, and each actuator force is bounded by
\(
F_j \in [F_{\mathrm{min}},F_{\mathrm{max}}]\ \mathrm{lb},
\)
where negative values represent inward forces. Given an actuator-force vector, the surrogate model predicts the fuselage displacement response, and the dimensional gap is evaluated at \(N=177\) measurement nodes along the fuselage cross-section.

We compare Q-Safe BO, C-Safe BO, and BO-ACL \citep{li2026bayesian} under two observation-noise levels, \(\sigma \in \{0.1,0.2\}\). These noise levels reflect practical measurement uncertainty in fuselage assembly. For a fair comparison, all methods are evaluated under the same response-estimation requirement: the dimensional-gap estimate must satisfy an accuracy tolerance of \(\epsilon_{\max}=0.04\) with \(95\%\) confidence. Since the measurement noise can be larger than this tolerance, repeated queries are required to obtain reliable response estimates before updating the optimization model.

It should be noted that this comparison is a sample-efficiency comparison rather than a wall-clock runtime comparison. In fuselage assembly, each oracle evaluation corresponds to a costly experiment: a physical shape-control trial on the assembly line or a high-fidelity FEA or surrogate-model simulation. The practical bottleneck is therefore the number of such evaluations required, rather than the processing time of each one. Accordingly, Q-Safe BO does not claim to reduce the runtime of an individual FEA or surrogate call. Its advantage is that it requires fewer oracle evaluations to estimate the mean response to a prescribed accuracy, and thus identifies high-quality feasible actuator-force configurations under a smaller evaluation budget. In practical manufacturing settings, reducing the number of physical trials or expensive simulations directly reduces experimental cost, equipment occupancy, and production disruption, which are savings that go well beyond reducing running time.

\subsection{Fuselage Assembly Results}
\subsubsection{Discrete Case}
\begin{figure}[!ht]
\centering
\includegraphics[width=0.5\linewidth]{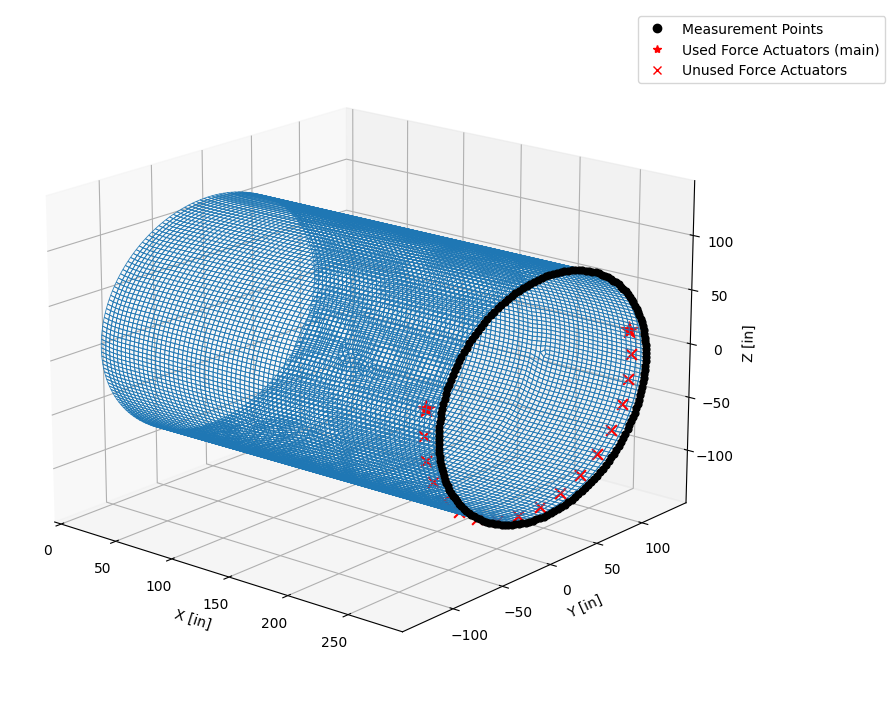}
\caption{Scheme of the discrete case. Stars: locations with active actuators.}
\label{fig:2actuators}
\end{figure}
In this subsection, we evaluate the proposed Q-Safe BO and classical baselines on fuselage assembly. To create a controlled setting with a known global optimum as a proof-of-concept, only two actuators are assigned active forces, while the remaining actuators are fixed at zero, as illustrated in Fig.~\ref{fig:2actuators}. For each active actuator, the feasible force range is discretized into 21 uniformly spaced values within \([-1000,1000]\) lb. Therefore, the discrete search space contains \(21^2=441\) candidate force configurations. Among these candidates, the global minimum MAE is \(0.072\) inches.

We compare four methods: Q-Safe BO on the simulator, Q-Safe BO (Real), C-Safe BO, and BO-ACL. Each method is evaluated under two observation noise levels, \(\sigma=0.1\) and \(\sigma=0.2\). For each method and noise level, we conduct five independent trials with different initial points.

\begin{figure}[!ht]
\centering
\begin{adjustbox}{max width=\textwidth}
    \includegraphics{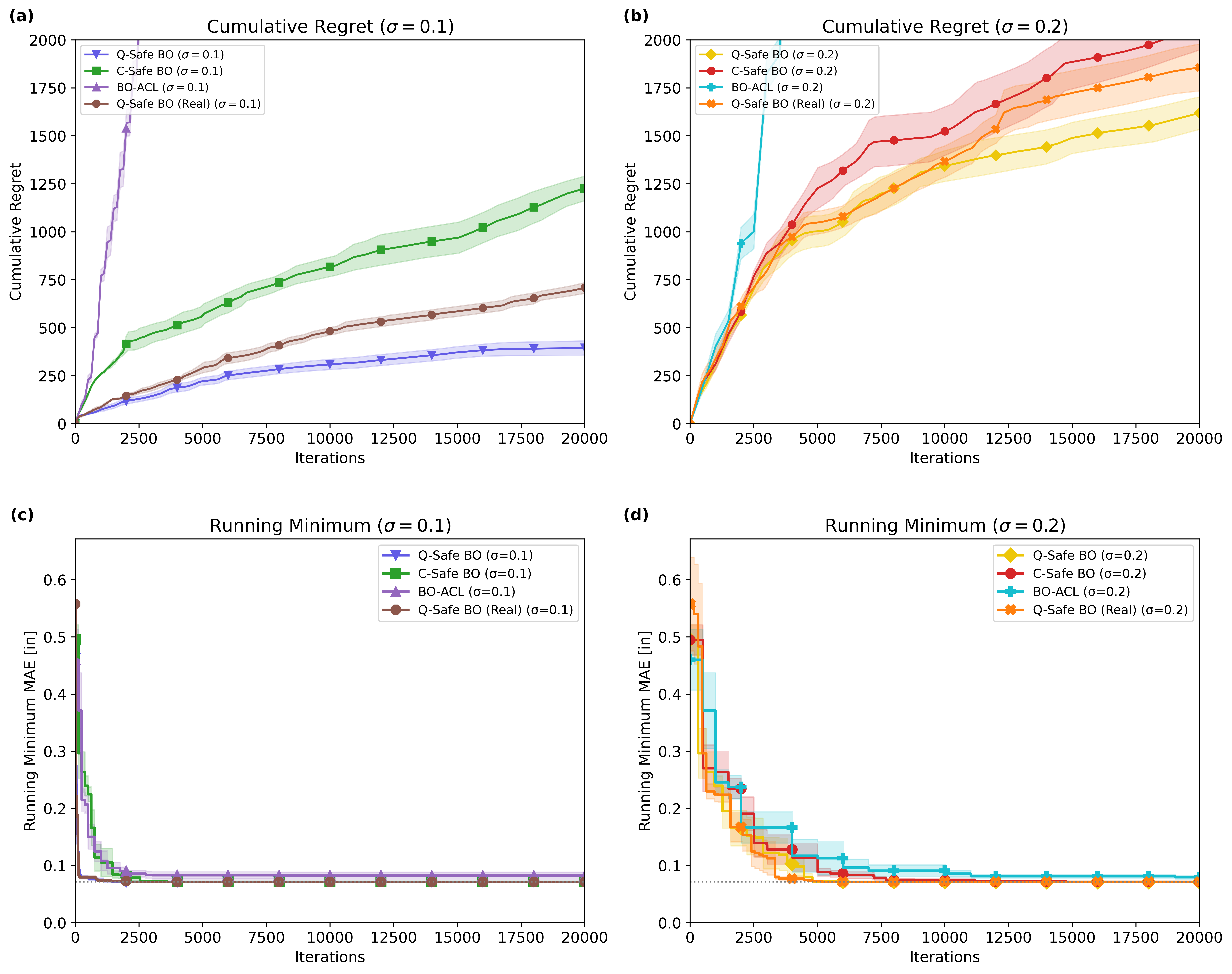}
\end{adjustbox}
\caption{Discrete two-actuator fuselage experiment. (a) and (b) Cumulative regret comparison under two observation noise levels. (c) Running minimum MAE under \(\sigma=0.1\). (d) Running minimum MAE under \(\sigma=0.2\).}
\label{fig:discumulative}
\end{figure}

Fig.~\ref{fig:discumulative}(a) and (b) present the cumulative regret curves for all methods under two noise levels. Q-Safe BO methods achieve lower cumulative regret than the classical baselines. In particular, Q-Safe BO achieves the lowest cumulative regret under \(\sigma=0.1\), while Q-Safe BO (Real) remains competitive and substantially outperforms C-Safe BO under the same noise level. When the observation noise increases to \(\sigma=0.2\), all methods receive larger regret, as expected. Nevertheless, Q-Safe BO methods still show better regret than C-Safe BO, indicating that QMC estimation uses the same oracle iterations more efficiently under noisy observations. Fig.~\ref{fig:discumulative}(c) and \ref{fig:discumulative}(d) further show the running minimum MAE under \(\sigma=0.1\) and \(\sigma=0.2\), respectively. These curves measure how quickly each method identifies a low dimensional gap force configuration. Under both noise levels, Q-Safe BO reaches the global minimum MAE of approximately \(0.072\) inches using fewer iterations than C-Safe BO. Specifically, when \(\sigma=0.1\), Q-Safe BO reaches the global optimum after an average of 800 iterations, whereas C-Safe BO requires 2,500 iterations. When the noise level increases to \(\sigma=0.2\), Q-Safe BO reaches the optimum after an average of 4,000 iterations, while C-Safe BO requires 8,000 iterations. Thus, Q-Safe BO consistently identifies the optimal force configuration faster than the C-Safe BO baseline.

Tab.~\ref{tab:discrete_violation_rate_noise} reports the constraint violation rates under the two observation noise levels. The violation rate is computed by strict evaluation using the Tsai--Wu surrogate model, rather than by using the GP-predicted feasibility labels. Across both noise levels, Q-Safe BO, Q-Safe BO (Real), and C-Safe BO maintain zero constraint violations, confirming that the nominal safe set consistently prevents the selection of unsafe force configurations. In contrast, BO-ACL violates the Tsai--Wu criterion under both noise levels. Its average violation rate increases from \(21.5\%\) when \(\sigma=0.1\) to \(31.5\%\) when \(\sigma=0.2\).

\begin{table}[!ht]
\centering
\scriptsize
\setlength{\tabcolsep}{3pt}
\renewcommand{\arraystretch}{1.05}

\begin{adjustbox}{max width=\columnwidth}
\begin{tabular}{lcccc}
\toprule
Method 
& \makecell{Avg. Viol. Rate\\\(\sigma=0.1\)} 
& \makecell{Unsafe Stages / Total\\\(\sigma=0.1\)} 
& \makecell{Avg. Viol. Rate\\\(\sigma=0.2\)} 
& \makecell{Unsafe Stages / Total\\\(\sigma=0.2\)} \\
\midrule
Q-Safe BO 
& \(0.000 \) 
& \(0 / 188\) 
& \(0.000 \) 
& \(0 / 153\) \\

Q-Safe BO (Real) 
& \(0.000 \) 
& \(0 / 299\) 
& \(0.000 \) 
& \(0 / 151\) \\

C-Safe BO 
& \(0.000 \) 
& \(0 / 156\) 
& \(0.000 \) 
& \(0 / 102\) \\

BO-ACL 
& \(0.215 \) 
& \(171 / 795\) 
& \(0.315 \) 
& \(63 / 200\) \\
\bottomrule
\end{tabular}
\end{adjustbox}
\caption{Constraint violation rates in the discrete fuselage experiment under two observation noise levels. Note that the ``Total'' counts BO stages, i.e., distinct actuator-force configurations selected, whereas the iteration counts in Fig.~\ref{fig:discumulative} and in the text refer to cumulative oracle queries. Each stage consumes multiple queries (see Sec.~\ref{sec:intro}).}
\label{tab:discrete_violation_rate_noise}
\end{table}

\subsubsection{Continuous Case}
In this subsection, we evaluate the constrained continuous actuator optimization problem. Each actuator has a continuous force range, and the optimization objective is to minimize the dimensional gap. In the continuous case, the true feasible optimum is unknown. We therefore compute regret with respect to the idealized reference $f^{*}=0$, which corresponds to a zero dimensional gap. This reference may not be achievable, but it provides a unified standard for comparing methods: since the MAE of the true feasible optimum satisfies $\mathbb{E}[\mathcal{L}(\mathbf{F}^{*})]\ge 0$, the reported cumulative regret upper-bounds the true regret, and the relative comparison among methods is unaffected.

\begin{figure}[ht]
\centering\includegraphics[width=1.0\linewidth]{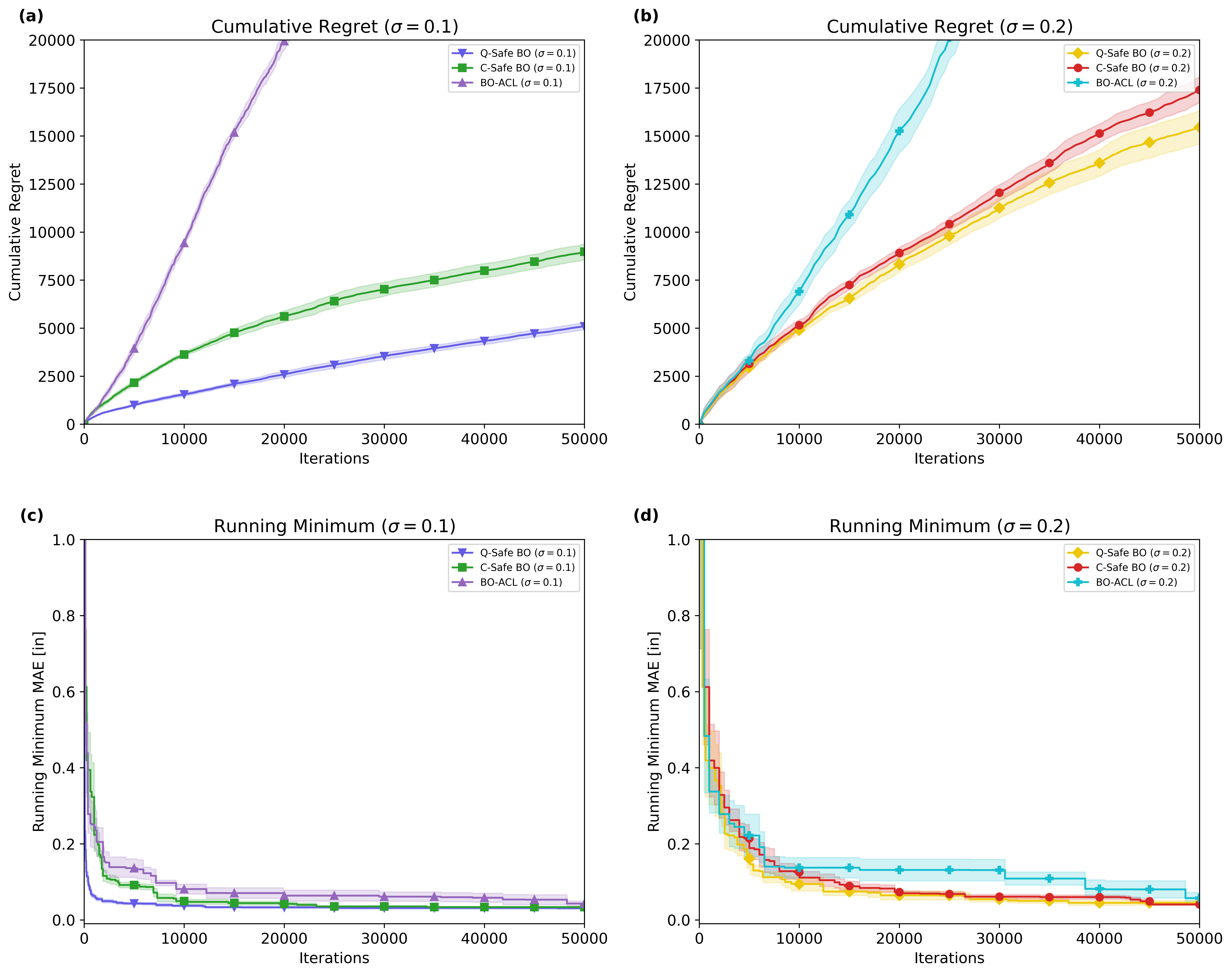}
\caption{(a) and (b) Comparison of cumulative regrets between Q-Safe BO, C-Safe BO, and BO-ACL using 8 actuators in fuselage assembly. (c) Running minimum MAE under \(\sigma=0.1\). (d) Running minimum MAE under \(\sigma=0.2\).}
\label{fig:cumulative}
\end{figure}

Fig.~\ref{fig:cumulative} (a) and (b) show cumulative regret, while (c) and (d) show the running minimum for each method with a specific initial dimensional gap under two noise levels. The cumulative regret curves indicate that Q-Safe BO rapidly identifies high-quality feasible actuator force configurations. The running minimum MAE curves further show that Q-Safe BO reaches a low dimensional gap within fewer iterations, especially under \(\sigma=0.1\).

An important advantage of Q-Safe BO and C-Safe BO is constraint satisfaction during optimization, shown in Tab.~\ref{tab:continuous_violation_rate_noise}. Both Q-Safe BO and C-Safe BO maintain a \(0\%\) violation rate under both noise levels. In contrast, BO-ACL incurs a \(55.9\%\) violation rate under \(\sigma=0.1\) and \(64.9\%\) under \(\sigma=0.2\). This result shows that although BO-ACL finds good individual solutions, it queries many infeasible or unsafe actuator configurations during the search process.

\begin{table}[ht]
\centering
\scriptsize
\setlength{\tabcolsep}{3pt}
\renewcommand{\arraystretch}{1.05}

\begin{adjustbox}{max width=\columnwidth}
\begin{tabular}{lcccc}
\toprule
Method 
& \makecell{Avg. Viol. Rate\\\(\sigma=0.1\)} 
& \makecell{Unsafe Stages / Total\\\(\sigma=0.1\)} 
& \makecell{Avg. Viol. Rate\\\(\sigma=0.2\)} 
& \makecell{Unsafe Stages / Total\\\(\sigma=0.2\)} \\
\midrule
Q-Safe BO 
& \(0.000 \) 
& \(0 / 1781\) 
& \(0.000 \) 
& \(0 / 778\) \\

C-Safe BO 
& \(0.000 \) 
& \(0 / 1169\) 
& \(0.000 \) 
& \(0 / 492\) \\

BO-ACL 
& \(0.559 \) 
& \(1670 / 2985\) 
& \(0.649 \) 
& \(976 / 1505\) \\
\bottomrule
\end{tabular}
\end{adjustbox}
\caption{Constraint violation rates in the continuous fuselage experiment under two observation noise levels across 5 trials for one initial condition.}
\label{tab:continuous_violation_rate_noise}
\end{table}

To further evaluate robustness and generalizability, we perform experiments over 10 distinct initial conditions. For each initial condition, each method is repeated for 5 trials with different initial points. Across all experiments, Q-Safe BO achieves the lowest aggregate mean minimum MAE under both noise levels, as reported in Tab.~\ref{tab:summary}.

\begin{table}[!ht]
\centering
\scriptsize
\begin{tabular}{l|cc}
\hline
\textbf{Method} & \textbf{\(\sigma=0.1\)} & \textbf{\(\sigma=0.2\)}\\
\hline
Q-Safe BO  & \textbf{0.027} \(\pm\) \textbf{0.009} & \textbf{0.041} \(\pm\) \textbf{0.012} \\

C-Safe BO  & 0.032 \(\pm\) 0.009 & 0.046 \(\pm\) 0.017 \\

BO-ACL     & 0.046 \(\pm\) 0.013 & 0.058 \(\pm\) 0.021 \\

\hline
\end{tabular}
\caption{Aggregate mean minimum MAE and standard deviation over 10 initial conditions. Lower values are better.}
\label{tab:summary}
\end{table}

\begin{figure}[ht]
  \centering
    \includegraphics[width=\textwidth]{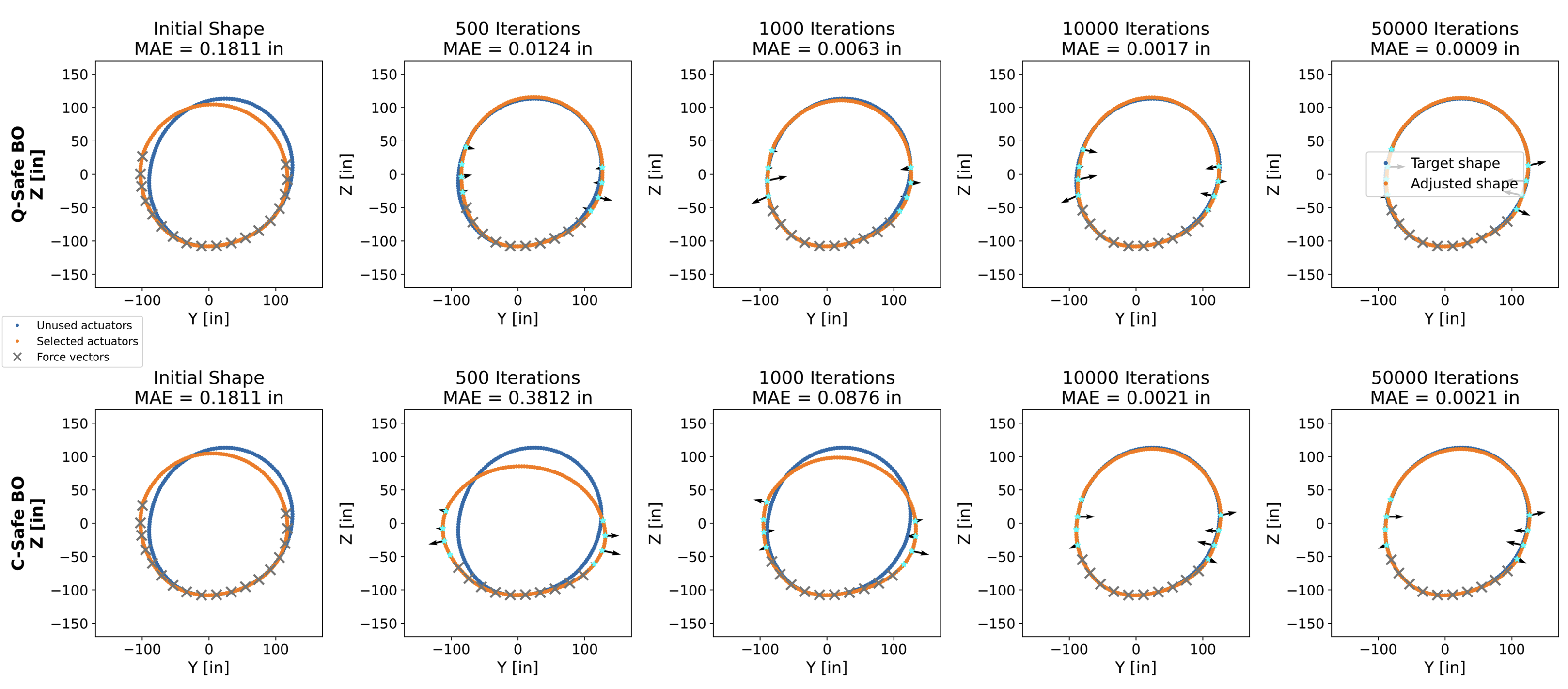}
  \caption{Shape adjustment results (cross section) using Q-Safe BO (Top) and C-Safe BO (Bottom) at noise level $\sigma = 0.1$. The length of each force vector represents the magnitude of the applied force. Both scenarios use the same initial condition. The best shape control results were obtained by C-Safe BO and Q-Safe BO at iterations 0 (Initial), 500, 1000, 10,000, and 50,000, respectively. MAE in this plot is calculated through the MAPDL software.}
  \label{fig:shape_adjustment}
\end{figure}

Fig.~\ref{fig:shape_adjustment} shows the cross-sectional shape adjustment results for Q-Safe BO and C-Safe BO under the same initial condition and noise level $\sigma=0.1$. The top row corresponds to Q-Safe BO, and the bottom row corresponds to C-Safe BO. Each column shows the shape control result obtained up to a specific iteration count: the initial shape, 500 iterations, 1000 iterations, 10,000 iterations, and 50,000 iterations. The orange curve represents the target shape, and the blue curve represents the adjusted shape. The arrows indicate the applied actuator-force vectors, where the vector length represents the force magnitude.

\begin{figure}[!ht]
\centering
\includegraphics[width=1.0\linewidth]{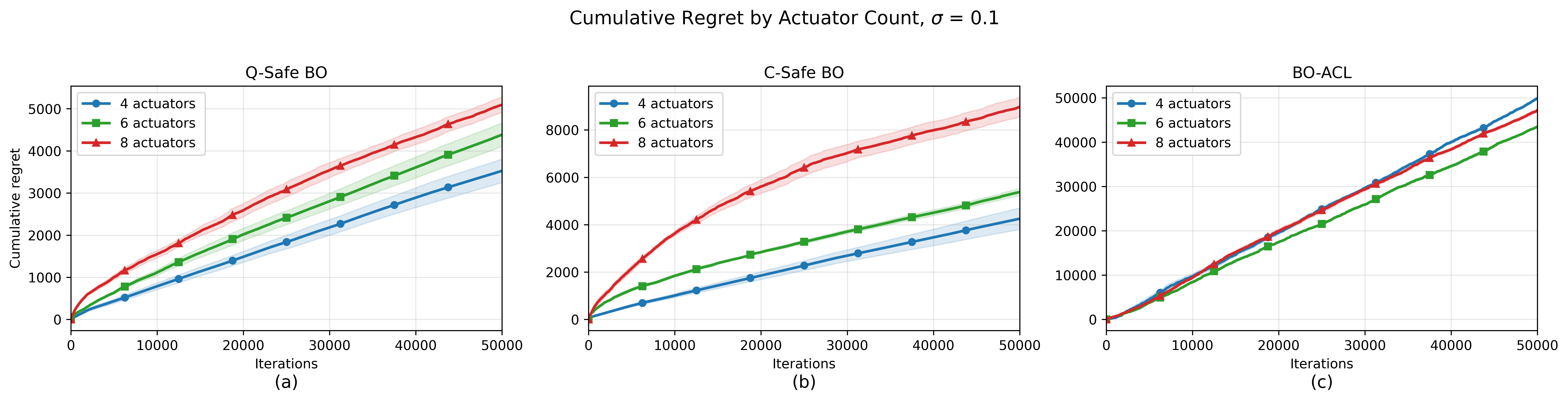}
\caption{Cumulative regret comparison under different actuator counts at noise level \(\sigma=0.1\). The results are shown for (a) Q-Safe BO, (b) C-Safe BO, and (c) BO-ACL with 4, 6, and 8 actuators, respectively.}
\label{fig:actuator_count_regret}
\end{figure}

To further investigate the sensitivity of cumulative regret to the problem scale, we examine two factors: the number of active actuators and the actuator-force range. Fig.~\ref{fig:actuator_count_regret} compares the cumulative regret of Q-Safe BO, C-Safe BO, and BO-ACL under 4, 6, and 8 active actuators with noise level \(\sigma=0.1\). For both Q-Safe BO and C-Safe BO, the cumulative regret increases as the number of actuators increases. This trend is expected because adding more actuators enlarges the continuous search space and increases the difficulty of identifying feasible high-quality force combinations.

\begin{figure}[ht]
\centering
\includegraphics[width=1.0\linewidth]{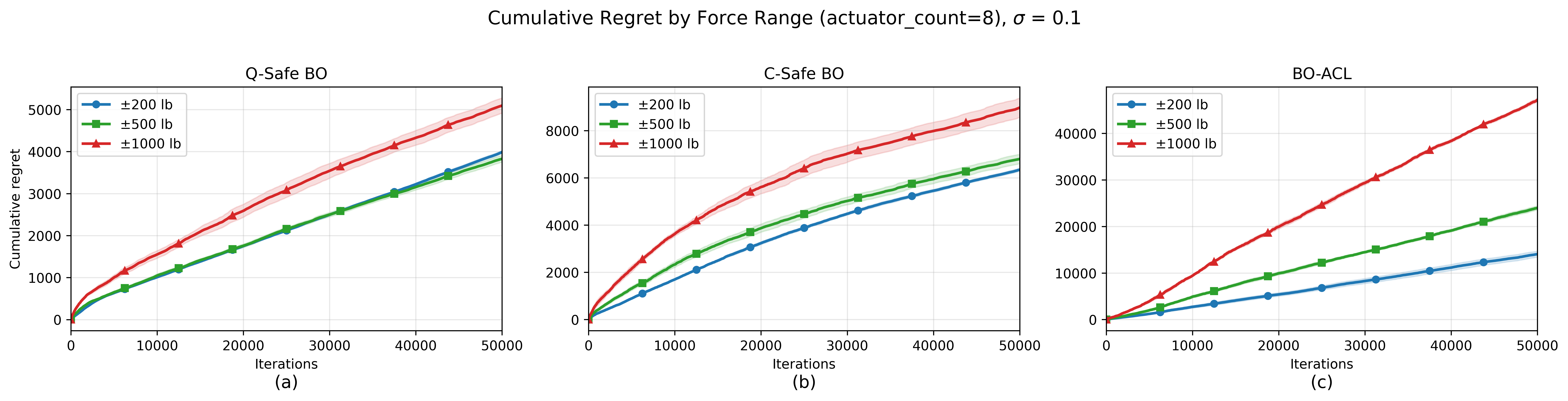}
\caption{Cumulative regret comparison under different actuator-force ranges with 8 active actuators and noise level \(\sigma=0.1\). As the force range expands from \(\pm 200\) lb to \(\pm 1000\) lb, cumulative regret generally increases, indicating that a larger search domain makes the constrained optimization problem more difficult. Q-Safe BO (a) achieves the lowest regret across all force ranges, while BO-ACL (c) shows the strongest sensitivity to the enlarged force range.}
\label{fig:force_range_regret}
\end{figure}

We also study the effect of the actuator-force range when the number of active actuators is fixed at 8. Fig.~\ref{fig:force_range_regret} compares the cumulative regret of Q-Safe BO, C-Safe BO, and BO-ACL under different actuator-force ranges with eight active actuators and noise level \(\sigma=0.1\). The force range is varied among \(\pm 200\) lb, \(\pm 500\) lb, and \(\pm 1000\) lb. Increasing the force range generally leads to higher cumulative regret, especially under the \(\pm 1000\) lb setting. For Q-Safe BO, the \(\pm 200\) lb and \(\pm 500\) lb cases show similar regret growth, while the \(\pm 1000\) lb case has noticeably larger cumulative regret. C-Safe BO shows a similar trend, but its cumulative regret is higher than that of Q-Safe BO across all force ranges. BO-ACL exhibits the largest sensitivity to the force range. In particular, when the force range increases to \(\pm 1000\) lb, its cumulative regret grows much more rapidly than in the smaller-range cases. Overall, Q-Safe BO achieves the lowest cumulative regret among the three methods across all tested force ranges.

\subsection{Discussion}
The experimental results provide several insights into the effectiveness of the proposed Q-Safe BO framework for fuselage shape control. Overall, the results show that Q-Safe BO improves sample efficiency, maintains safety during optimization, and performs robustly under different noise levels, actuator dimensions, and actuator-force ranges.

First, the comparison between Q-Safe BO and C-Safe BO demonstrates the benefit of QMC estimation under noisy observations. In experiments, all methods are required to satisfy the same estimation-accuracy requirement, \(\epsilon_{\max}=0.04\), with a \(95\%\) confidence level for the response. The classical baselines satisfy this requirement through repeated noisy evaluations by MC, whereas Q-Safe BO uses QMC estimation. Compared with MC, QMC estimation uses the available oracle budget more efficiently. Under the same query budget, Q-Safe BO can obtain more accurate objective estimates, update the surrogate model more effectively, and identify promising actuator-force configurations faster than the classical baseline. This explains why Q-Safe BO reaches the global minimum MAE faster in the discrete case and achieves lower aggregate minimum MAE in the continuous case.

Second, the results show that observation noise directly affects optimization performance. When the noise level increases from \(\sigma=0.1\) to \(\sigma=0.2\), all methods require more iterations to identify high-quality force configurations and incur larger regret. This is expected because larger noise magnitude makes the true dimensional gap harder to estimate. Nevertheless, Q-Safe BO remains better than or competitive with the classical baselines under both noise levels. The experimental results are consistent with the bounded variance regret analysis in Theorem~\ref{thm:modified_bounded_variance_regret_paper}, which shows that the regret bound depends explicitly on the noise magnitude \(\sigma\). Therefore, the larger regret observed at \(\sigma=0.2\) is theoretically expected.

Third, the constraint violation results show that explicitly modeling safety is necessary for manufacturing optimization. In both the discrete and continuous experiments, Q-Safe BO and C-Safe BO maintain zero constraint violations under the Tsai--Wu failure criterion. In contrast, BO-ACL queries many unsafe actuator-force configurations. This difference shows that a method may occasionally identify good objective values, but it is still unsuitable for safety-critical manufacturing if it frequently evaluates infeasible or unsafe configurations. By restricting the search to the nominal safe region, Q-Safe BO improves objective performance while preserving structural feasibility throughout the optimization process.

Finally, the real IBM hardware experiment provides an initial validation of the deployability of the proposed quantum-enhanced framework. Although the hardware experiment is conducted on the discrete two-actuator setting due to current quantum hardware limitations, Q-Safe BO (Real) remains competitive with simulator-based Q-Safe BO and outperforms C-Safe BO under two noise levels. This result suggests that the proposed framework is not limited to idealized simulation. However, current quantum hardware still has limitations, including hardware noise and restrictions on circuit depth. Therefore, Q-Safe BO should not be interpreted as universally superior to classical BO in all settings. Instead, theoretical and empirical results suggest that Q-Safe BO is most promising when objective evaluations are noisy and expensive and the effective actuator dimension is moderate.

\section{Conclusion}\label{sec:conclusion}
In this paper, we propose a Q-Safe BO framework for optimal actuator-force design in fuselage assembly. In aircraft
manufacturing, fuselage sections are manufactured separately and later joined
together. During this joining process, a dimensional gap may occur between
adjacent sections, which can compromise assembly quality and structural safety.
Force-controlled actuators are commonly used to adjust the fuselage shape and reduce the dimensional gap. The fuselage assembly problem presents three main challenges. First, each candidate actuator-force configuration requires an expensive evaluation of the resulting dimensional gap. Second, measurement noise makes a single observation unreliable, so repeated measurements are needed to obtain an accurate estimate of the true response. Third, the selected forces must satisfy structural safety constraints to prevent material failure, excessive residual stress, or other damage during assembly.

To address these challenges, the proposed framework combines BO, QMC estimation, safe-set exploration, and surrogate modeling. The objective is to minimize the dimensional gap, while ensuring that each selected actuator force configuration remains structurally feasible. Specifically, we define a safety requirement based on the Tsai--Wu failure criterion, where a force configuration is considered safe if the predicted failure index satisfies \(FI\leq1\). During optimization, the model is restricted to the safe region so that it prioritizes actuator force configurations that are likely to improve shape control while maintaining structural feasibility. In this way, Q-Safe BO explicitly balances optimization performance and safety during the search process.

Compared with the classical baselines, the proposed framework leverages QMC estimation to improve query efficiency under noisy observations. Rather than reducing the wall-clock cost of each individual structural evaluation, the main advantage of Q-Safe BO is that it reduces the number of oracle queries required to estimate the expected dimensional gap to a prescribed accuracy. This advantage is especially relevant in fuselage assembly, where repeated FEA or surrogate-model evaluations may be required to obtain reliable estimates under measurement noise. The bounded variance regret analysis in Appendix~\ref{thm:modified_bounded_variance_regret_appdix} further shows that the regret depends directly on the noise magnitude $\sigma$ and the effective actuator dimension $d$.

The experimental results support this interpretation. In the discrete two-actuator case, Q-Safe BO reaches the global minimum MAE of approximately 0.072 inches using fewer iterations than C-Safe BO under both noise levels. The real IBM hardware experiment further demonstrates that the proposed framework can be executed on current quantum hardware, although the search space is restricted due to present hardware limitations. In the continuous eight-actuator case, Q-Safe BO achieves lower aggregate minimum MAE than C-Safe BO and BO-ACL under both $\sigma=0.1$ and $\sigma=0.2$. Moreover, Q-Safe BO and C-Safe BO maintain zero constraint violations in both the discrete and continuous experiments, while BO-ACL queries a substantial number of unsafe force configurations. These results show that the proposed method improves sample efficiency and solution quality while preserving structural safety during the optimization process.

The actuator-count and force-range studies further confirm the theoretical discussion. As the number of active actuators increases from 4 to 8, the cumulative regret increases for Q-Safe BO and C-Safe BO because the continuous search space becomes larger and more difficult to explore. A similar trend is observed when the actuator-force range expands from \(\pm 200\) lb to \(\pm 1000\) lb. A wider force range enlarges the feasible search domain and introduces more extreme force combinations, which makes it more challenging to identify high-quality safe configurations. Nevertheless, Q-Safe BO consistently maintains lower cumulative regret than C-Safe BO and BO-ACL across the tested actuator counts and force ranges. This suggests that the quantum-enhanced mean-estimation step improves the efficiency of safe exploration, particularly when the optimization problem involves noisy observations, increasing actuator-force dimension, and a larger continuous force domain.

Although the proposed framework achieves promising results, several directions remain for future work. First, the current study considers the Tsai--Wu failure criterion as the primary safety requirement. In real fuselage assembly, additional physical constraints may need to be incorporated, such as residual stress, actuator saturation, local deformation limits, fixture accessibility, and assembly-process uncertainty. Extending the safe set model to handle multiple coupled constraints would further improve the practical applicability of the method. Second, the current quantum oracle and mean estimation are abstract subroutines. In future work, we plan to develop a more complete quantum learning pipeline by constructing a real quantum prediction model combined with MC or QMC estimation. Such a model would allow the surrogate prediction and uncertainty estimation to be generated by an explicit quantum architecture, rather than treating QMC as an abstract subroutine. This extension would provide a more realistic path toward hardware-level QBO for manufacturing applications.

\newpage
\appendix

\section{Bayesian Optimization}\label{appdix:clsBO}
We briefly introduce vanilla BO. For a complete introduction to BO, the reader may refer to \cite{frazier2018bayesian, garnett2023bayesian, srinivas2009gaussian}. BO is a powerful black-box optimization technique designed to efficiently identify optimal solutions to expensive-to-evaluate objective functions. Consider the following optimization problem: 
\begin{equation*}
    y = f(x) + \xi,
\end{equation*} where $y$ represents the noisy observations with Gaussian noise $\xi$, and $f(\cdot)$ is an expensive-to-evaluate black-box function. The goal is to find the optimal solution $x_{*}$ such that:
\begin{equation*}
    x_{*} = \argmax_{x \in \mathcal{A}} f(x),
\end{equation*} where $\mathcal{A}$ denotes the feasible set. BO adopts a Sequential Model-Based Optimization (SMBO) framework, which primarily consists of two crucial components: (i) surrogate model, and (ii) optimization strategy.

The idea of the surrogate model is to model the response from the environment with respect to different inputs. The most common surrogate model within the BO framework is the Gaussian Process (GP). In GP, we assume that the objective function $f(x)$ follows a Gaussian Process such that: $f(x) \sim GP(m(x), K(x,x^{\prime}))$, where $m(x)$ is the mean function and $K(x,x^{\prime})$ is the covariance function. The mean function is the expectation of the objective function $m(x) = \mathbb{E}[f(x)]$, while the covariance function (kernel) describes the difference between outputs among different inputs $K(x,x^{\prime}) = \mathbb{E}[(f(x) - m(x))(f(x^{\prime}) - m(x^{\prime}))]$. Several kernels can be employed, such as the linear kernel, Radial Basis Function (RBF) kernel, and Matérn kernel. In this paper, we adopt the RBF kernel: $K(\mathbf{x}, \mathbf{x}') = \exp\left( -\frac{\|\mathbf{x} - \mathbf{x}'\|^2}{2l^2} \right)$.

After modeling the response from the environment using GP, we can predict the output using the posterior of GP. Given historical observations $\mathcal{D}_{t} = \{\mathbf{X}_{t}, \mathbf{y}_{t}\}$, the posterior mean and uncertainty at the new input $x_{*}$ are expressed as:
\begin{equation}\label{eq:posterior}
\begin{aligned}
m_t(x_{*}) &= K(x_{*}, \mathbf{X}_t) \left[ K(\mathbf{X}_t, \mathbf{X}_t) + \sigma^2 \mathbf{I} \right]^{-1} \mathbf{y}_t \\
\sigma_{t}^{2}(x_{*}) &= K(x_{*}, x_{*}) - K(x_{*}, \mathbf{X}_t) \left[ K(\mathbf{X}_t, \mathbf{X}_t) + \sigma^2 \mathbf{I} \right]^{-1} K(\mathbf{X}_t, x_{*})
\end{aligned}
\end{equation}

For the optimization strategy, the essential step is selecting an appropriate acquisition function that balances exploration and exploitation when determining the next point to evaluate. In this paper, we utilize Upper Confidence Bound as an acquisition function, defined in Eq. (\ref{eq:UCB}):

\begin{equation}\label{eq:UCB}
    x_{*} = \argmax_{x \in \mathcal{A}} m_{t}(x) + \beta \sigma_{t}(x),
\end{equation}
where $\beta$ controls the trade-off between exploration (favoring points with high uncertainty) and exploitation (favoring points with high predictive mean). Initially, when limited samples are available, UCB prioritizes exploration. As more data is collected, the algorithm increasingly exploits points with higher predicted means.

\section{Proof of Theorem 1}\label{Appdix:B}

\noindent\textbf{Notation.} Throughout Appendices~\ref{Appdix:B} and~\ref{Appdix:c}, we drop the tilde
for readability: $\mu_{f,s-1}(\cdot)$ and $\sigma_{f,s-1}(\cdot)$ denote the
\emph{weighted} GP posterior mean and standard deviation of the objective
model, written as $\tilde{\mu}_{f,s-1}(\cdot)$ and
$\tilde{\sigma}_{f,s-1}(\cdot)$ in the main text (Eqs.~\eqref{eq:objective_surrogate}).

\begin{lemma}
Let $f \in \mathcal H_k$ satisfy $\|f\|_k \le B$, and assume the kernel is bounded on the diagonal:
\[
\sup_{x\in\mathcal D} k(x,x)\le \alpha^2 <\infty.
\]
Suppose the observations are
\[
y_i = f(x_i)+\xi_i,\qquad i=1,\dots,t,
\]
where the noises satisfy
\[
\mathbb E[\xi_i]=0,
\qquad
\operatorname{Var}(\xi_i)\le \sigma^2<\infty.
\]
Let
\[
\mu_t(x)=k_t(x)^\top (K_t+\lambda I)^{-1} y_{1:t},
\qquad \lambda>0,
\]
be the GP posterior mean. Then for every $x\in\mathcal D$:
$\mu_t(x)$ is finite.
\end{lemma}
\begin{proof}
Write
\[
y_{1:t}=f_{1:t}+\xi_{1:t},
\]
where
\[
f_{1:t}=(f(x_1),\dots,f(x_t))^\top.
\]
Then
\[
\mu_t(x)
=
k_t(x)^\top (K_t+\lambda I)^{-1} f_{1:t}
+
k_t(x)^\top (K_t+\lambda I)^{-1} \xi_{1:t}.
\]
Since $K_t+\lambda I$ is positive definite for $\lambda>0$, its inverse exists and has finite entries. Also, each entry of $k_t(x)$ is finite because
\[
|k(x_i,x)|\le \sqrt{k(x_i,x_i)k(x,x)}\le \alpha^2
\] by the Cauchy-Schwarz inequality. 
Further, since $f\in\mathcal H_k$ and $\|f\|_k\le B$, by the reproducing property,
\[
|f(x_i)|\le \|f\|_k\sqrt{k(x_i,x_i)}
,\]
so $f_{1:t}$ has finite entries. Therefore $\mu_t(x)$ is a finite linear combination with two finite terms. Hence, $\mu_t$ is finite. 
\end{proof}

\paragraph{Modified precision rule.}
We establish the maximum precision for the estimation. We restate the theorems from \cite{dai2023quantum} as necessary. Pick a constant $c \in (0,1]$. We consider the variant of Q-GP-UCB in which
\[
\epsilon_s^{(c)} \triangleq c\,\frac{\widetilde{\sigma}_{s-1}^{(c)}(x_s)}{\sqrt{\lambda}} \le \epsilon_{\max}, 
\qquad
w_s^{(c)} \triangleq \frac{1}{(\epsilon_s^{(c)})^2}.
\]
Define
\[
V_s^{(c)}
\triangleq
\lambda I + \sum_{\tau=1}^s \frac{1}{(\epsilon_\tau^{(c)})^2}\phi(x_\tau)\phi(x_\tau)^\top,
\]
and
\[
\widetilde{\sigma}_s^{(c)}(x)^2
\triangleq
\lambda\,\phi(x)^\top (V_s^{(c)})^{-1}\phi(x).
\]
Also define the weighted information gain
\[
\widetilde{\gamma}_s^{(c)}
\triangleq
\frac12 \log\det\!\Bigl(I+\lambda^{-1}\widetilde{K}_s^{(c)}\Bigr),
\]
where $\widetilde{K}_s^{(c)}$ is the weighted kernel matrix induced by
$w_1^{(c)},\dots,w_s^{(c)}$.

\begin{lemma}[Weighted information-gain order under modified precision]
\label{lemma3}
Assume the same QMC estimation budget model as in \cite{dai2023quantum}, with
\[
N_{\epsilon^{(c)}}
=
\frac{C_1}{\epsilon^{(c)}}\log\frac{2\bar m^{(c)}}{\delta}
\qquad
\text{(bounded noise)}
\], where $\bar m^{(c)}$ is an upper bound on total stages.
Then, following Lemma 2 of \citet{dai2023quantum} and choosing $\delta \in (0, 2/e]$, we have
\[
\sum_{s=1}^{m^{(c)}} \frac{1}{(\epsilon_s^{(c)})^2}\le T^{2}
\]
Consequently, the analogue of Theorem 1 and Corollary 1 \citep{dai2023quantum} implies
\[
\widetilde{\gamma}_{m^{(c)}}^{(c)} = O(d\log T)
\quad\text{for the linear kernel,}
\]
and
\[
\widetilde{\gamma}_{m^{(c)}}^{(c)} = O((\log T)^{d+1})
\quad\text{for the SE kernel.}
\]
\end{lemma}

\begin{proof}
Let \(N_s^{(c)} := N_s\),
which is the number of queries used at stage $s$. 
Let $\bar m^{(c)}$ be an upper bound on the number of stages,
so that \(m^{(c)}\le \bar m^{(c)}\). By the total QMC estimation budget constraint,
\[
\sum_{s=1}^{m^{(c)}} N_s^{(c)} \le T.
\]
First consider the bounded-noise case. Since
\[
N_s
=
\frac{C_1}{\epsilon_s^{(c)}}
\log\frac{2\bar m}{\delta},
\]
we have
\[
\sum_{s=1}^{m^{(c)}} \frac{1}{\epsilon_s^{(c)}}
=
\frac{1}{C_1\log(2\bar m/\delta)}
\sum_{s=1}^{m^{(c)}} N_s^{(c)}
\le
\frac{T}{C_1\log(2\bar m/\delta)}
=
\mathcal{O}(T).
\]
Therefore, since all terms are nonnegative,
\[
\sum_{s=1}^{m^{(c)}}\frac{1}{(\epsilon_s^{(c)})^2}
\le
\left(
\sum_{s=1}^{m^{(c)}}\frac{1}{\epsilon_s^{(c)}}
\right)^2
=
\mathcal{O}(T^2).
\]
Now define
\[
S_c :=
\sum_{s=1}^{m^{(c)}}\frac{1}{(\epsilon_s^{(c)})^2}.
\]
The information-gain bounds in the analogue of Theorem 1 and Corollary 1 \citep{dai2023quantum} depend on the accuracy sequence only through $S_c$. Since $S_c=\mathcal{O}(T^2)$, the linear-kernel bound becomes
\[
\widetilde{\gamma}_{m^{(c)}}^{(c)}
=
\mathcal{O}\left(d\log(S_c)\right)
=
\mathcal{O}(d\log T),
\]
and the SE-kernel bound becomes
\[
\widetilde{\gamma}_{m^{(c)}}^{(c)}
=
\mathcal{O}\left((\log(S_c))^{d+1}\right)
=
\mathcal{O}((\log T)^{d+1}).
\]
\end{proof}

\begin{lemma}
Suppose that there exist constants \(B_g>0\) and \(\sigma_{\min}>0\) such that
\[
|\mu_{g,s-1}(x)|\le B_g,
\qquad
\sigma_{g,s-1}(x)\ge \sigma_{\min},
\qquad
\forall x\in \widehat S_s,\ \forall s.
\]
Let
\[
z_s(x)=\frac{\mu_{g,s-1}(x)}{\sigma_{g,s-1}(x)}.
\]
Then there exists a constant \(M_z>0\) such that
\[
|z_s(x)|\le M_z,
\qquad
\forall x\in \widehat S_s,\ \forall s.
\]
In particular, one may take \(M_z=B_g/\sigma_{\min}\).
\end{lemma}

\begin{proof}
For any \(x\in \widehat S_s\), we have
\[
|z_s(x)|
=
\left|\frac{\mu_{g,s-1}(x)}{\sigma_{g,s-1}(x)}\right|
=
\frac{|\mu_{g,s-1}(x)|}{\sigma_{g,s-1}(x)}.
\]
Using the assumptions
\[
|\mu_{g,s-1}(x)|\le B_g
\qquad\text{and}\qquad
\sigma_{g,s-1}(x)\ge \sigma_{\min},
\]
we obtain
\[
|z_s(x)|
\le
\frac{B_g}{\sigma_{\min}}.
\]
Therefore, by setting
\[
M_z=\frac{B_g}{\sigma_{\min}},
\]
we have
\[
|z_s(x)|\le M_z,
\qquad
\forall x\in \widehat S_s,\ \forall s.
\]
This proves the claim.
\end{proof}

\begin{lemma}[Modified stage-count bound]
\label{lemma:stage-count}
For every $\tau = 0,\dots,m^{(c)}-1$,
\[
\det\!\bigl(V_{\tau+1}^{(c)}\bigr)
=
\Bigl(1+\frac{1}{c^2}\Bigr)\det\!\bigl(V_\tau^{(c)}\bigr).
\]
Hence,
\[
m^{(c)} \log\!\Bigl(1+\frac{1}{c^2}\Bigr)
=
\log\frac{\det(V_{m^{(c)}}^{(c)})}{\det(V_0^{(c)})}.
\]
Moreover, using
\[
\log\frac{\det(V_{m^{(c)}}^{(c)})}{\det(V_0^{(c)})}
=
2\widetilde{\gamma}_{m^{(c)}}^{(c)},
\]
we obtain
\[
m^{(c)}
\le
\frac{2\widetilde{\gamma}_{m^{(c)}}^{(c)}}{\log(1+c^{-2})}.
\]
Therefore,
\[
m^{(c)} = \mathcal{O}\Bigl(\frac{d\log T}{\log(1+c^{-2})}\Bigr)
\quad\text{for the linear kernel,}
\]
and
\[
m^{(c)} = \mathcal{O}\Bigl(\frac{(\log T)^{d+1}}{\log(1+c^{-2})}\Bigr)
\quad\text{for the SE kernel.}
\]
\end{lemma}

\begin{proof}
From proof of Theorem 2 \citep{dai2023quantum}, we have the following:
\begin{align*}
\det(V_{\tau+1}^{(c)})
&=
\det
\left(
\lambda I + \sum_{j=1}^{\tau + 1} \frac{1}{(\epsilon_{j}^{(c)})^2}
\phi(x_{j})\phi(x_{j})^\top
\right)\\
&=
\det\left(\lambda I + \sum_{j=1}^{\tau} \frac{1}{(\epsilon_{j}^{(c)})^2} \phi(x_j)\phi(x_j)^\top + \frac{1}{(\epsilon_{\tau+1}^{(c)})^2}\phi(x_{\tau+1})\phi(x_{\tau+1})^\top\right)
\\
&=
\det\left(V_\tau^{(c)} + \frac{1}{(\epsilon_{\tau+1}^{(c)})^2}\phi(x_{\tau+1})\phi(x_{\tau+1})^\top\right)\\
&= \det\left({V_\tau^{(c)}}^{1/2} \left(I + \frac{1}{(\epsilon_{\tau+1}^{(c)})^2} {V_\tau^{(c)}}^{-1/2} \phi(x_{\tau+1})\phi(x_{\tau+1})^\top {V_\tau^{(c)}}^{-1/2} \right) {V_\tau^{(c)}}^{1/2}\right) \\
&= \det(V_\tau^{(c)}) \det\left(I + \frac{1}{(\epsilon_{\tau+1}^{(c)})^2} {V_\tau^{(c)}}^{-1/2} \phi(x_{\tau+1})\phi(x_{\tau+1})^\top {V_\tau^{(c)}}^{-1/2} \right) \\
&= \det(V_\tau^{(c)}) \left(1 + \frac{1}{(\epsilon_{\tau+1}^{(c)})^2} \phi(x_{\tau+1})^\top {V_\tau^{(c)}}^{-1/2} {V_\tau^{(c)}}^{-1/2} \phi(x_{\tau+1}) \right) \\
&= \det(V_\tau^{(c)}) \left(1 + \frac{1}{(\epsilon_{\tau+1}^{(c)})^2} \|\phi(x_{\tau+1})\|_{{V_\tau^{(c)}}^{-1}}^2 \right)
\end{align*}
Using
\[
\widetilde{\sigma}_\tau^{(c)}(x_{\tau+1})^2
=
\lambda\,\phi(x_{\tau+1})^\top (V_\tau^{(c)})^{-1}\phi(x_{\tau+1}),
\]
and
\[
(\epsilon_{\tau+1}^{(c)})^2
=
c^2\frac{\widetilde{\sigma}_\tau^{(c)}(x_{\tau+1})^2}{\lambda},
\]
we get
\[
\frac{1}{(\epsilon_{\tau+1}^{(c)})^2}
\phi(x_{\tau+1})^\top (V_\tau^{(c)})^{-1}\phi(x_{\tau+1})
=
\frac{1}{c^2}.
\]
Thus
\[
\det(V_{\tau+1}^{(c)})
=
\Bigl(1+\frac{1}{c^2}\Bigr)\det(V_\tau^{(c)}).
\]
Define $V_0^{(c)}=\lambda I$. Iterating the previous identity gives
\[
\det(V_{m^{(c)}}^{(c)})
=
(1+c^{-2})^{m^{(c)}}\det(V_0^{(c)}).
\]
By the weighted information-gain identity,
\[
m^{(c)}\log(1+c^{-2})=\log\frac{\det(V_{m^{(c)}}^{(c)})}{\det(V_0^{(c)})}
=
2\widetilde{\gamma}_{m^{(c)}}^{(c)}.
\] Define
\[
S_c
=
\sum_{s=1}^{m^{(c)}}\frac{1}{(\epsilon_s^{(c)})^2}.
\]
For the linear kernel, Corollary 1 \citep{dai2023quantum} and Lemma 3 give
\[
\widetilde{\gamma}_{m^{(c)}}^{(c)}
\le
Cd\log(S_c) = \mathcal{O}(d\log T).
\]
Hence
\[
m^{(c)}
\le
\frac{2Cd\log(S_c)}{\log(1+c^{-2})}=\mathcal{O}\Bigl(\frac{d\log T}{\log(1+c^{-2})}\Bigr).
\]

% For the linear kernel, the algorithm takes at most $\mathcal{O}(d\log T)$ stages.

On the other hand, for the SE kernel, there exists
a constant \(C_{\mathrm{SE}}>0\) such that
\[
\widetilde{\gamma}_{m^{(c)}}^{(c)}
\le
C_{\mathrm{SE}}
\bigl(\log(S_c)\bigr)^{d+1} = \mathcal{O}\bigl(\log(T)\bigr)^{d+1},
\]
where
\[
S_c
=
\sum_{s=1}^{m^{(c)}}\frac{1}{(\epsilon_s^{(c)})^2}.
\]
Combining this with the determinant-growth identity gives
\[
m^{(c)}\log(1+c^{-2})
=
\log
\frac{\det(V_{m^{(c)}}^{(c)})}{\det(V_0^{(c)})}
=
2\widetilde{\gamma}_{m^{(c)}}^{(c)}
\le
2C_{\mathrm{SE}}
\bigl(\log(S_c)\bigr)^{d+1}.
\]
Therefore,
\[
m^{(c)}
\le
\frac{2C_{\mathrm{SE}}}{\log(1+c^{-2})}
\bigl(\log(S_c)\bigr)^{d+1} = O\!\left(
\frac{(\log T)^{d+1}}{\log(1+c^{-2})}
\right).
\]

\end{proof}

\begin{theoremone}[Bounded noise regret of Q-Safe BO]
Let \(\mathcal X\) be compact. In stage \(s\), define the nominal current safe 
set by
\[
\widehat S_s
=
\{x\in\mathcal X:\mathrm{LCB}_{\hat{g},s}(x)\ge0\},
\]
where
\[
\mathrm{LCB}_{\hat{g},s}(x)
=
\mu_{g,s-1}(x)
-
\beta_s^{(\mathrm{con})}\sigma_{g,s-1}(x).
\]
Assume that each nominal safe set is nonempty:
\[
\widehat S_s\neq\emptyset,
\qquad
\forall s.
\]
For each stage \(s\), define the nominal-set oracle comparator by
\[
x_s^\dagger
\in
\arg\max_{x\in\widehat S_s} f(x).
\]
Thus, \(x_s^\dagger\) is the best point in the current nominal safe set.

Pick a constant \(c\in(0,1]\), and suppose the quantum estimation precision at
stage \(s\) is chosen as
\[
\epsilon_s^{(c)}
=
c\,\frac{\sigma_{f,s-1}(x_s)}{\sqrt{\lambda}}
\le \epsilon_{\max}.
\]

Assume the following conditions hold:
\begin{enumerate}
    \item The objective confidence event holds uniformly with probability at
    least \(1-\delta\). That is, on an event \(\mathcal E_f\) satisfying
    \[
    \mathbb P(\mathcal E_f)\ge1-\delta,
    \]
    we have
    \[
    |f(x)-\mu_{f,s-1}(x)|
    \le
    \beta_s^{(f,c)}\sigma_{f,s-1}(x),
    \qquad
    \forall x\in\mathcal X,\ \forall s.
    \]

    \item The selected point \(x_s\) satisfies
    \[
    x_s\in\arg\max_{x\in\widehat S_s} A_s(x),
    \]
    where
    \[
    A_s(x)
    =
    (1-\eta_s)\mathrm{UCB}_{f,s}(x)
    -
    \eta_s z_s(x),
    \]
    with
    \[
    \mathrm{UCB}_{f,s}(x)
    =
    \mu_{f,s-1}(x)
    +
    \beta_s^{(f,c)}\sigma_{f,s-1}(x),
    \]
    and
    \[
    z_s(x)
    =
    \left|
    \frac{\mu_{g,s-1}(x)}
    {\sigma_{g,s-1}(x)}
    \right|.
    \]

    \item There exists \(M_z>0\) such that
    \[
    |z_s(x)|\le M_z,
    \qquad
    \forall x\in\widehat S_s,\ \forall s.
    \]

    \item There exists \(\sigma_{\min}>0\) such that
    \[
    \sigma_{f,s-1}(x_s)\ge\sigma_{\min},
    \qquad
    \forall s.
    \]

    \item The mixing coefficient is chosen as
    \[
    \eta_s=\frac{1}{s+1}.
    \]

    \item The sequence \(\beta_s^{(f,c)}\) is nondecreasing in \(s\).

    \item At stage \(s\), the quantum subroutine uses
    \[
    N_s
    \le
    \frac{C_1}{\epsilon_s^{(c)}}
    \log\left(\frac{2\bar m^{(c)}}{\delta}\right)
    \]
    oracle queries.

    \item For the SE kernel, the number of stages satisfies
    \[
    m^{(c)}
    \le
    \bar m^{(c)}
    =
    \mathcal{O}\left(
    \frac{(\log T)^{d+1}}
    {\log(1+c^{-2})}
    \right),
    \]
    and the confidence width satisfies
    \[
    \beta_{m^{(c)}}^{(f,c)}
    = B + \sqrt{2\left(\tilde{\gamma}_{m^{(c)}-1}+1+\log(\frac{2}{\delta})\right)}
    =
    \mathcal{O}\left(
    (\log T)^{\frac{d+1}{2}}
    \right).
    \]
\end{enumerate}

Define the stage-wise nominal-set regret by
\[
\widetilde r_s
=
f(x_s^\dagger)-f(x_s).
\]
Since \(x_s^\dagger\) maximizes \(f\) over \(\widehat S_s\), and since
\(x_s\in\widehat S_s\), we have
\[
\widetilde r_s\ge0.
\]

Define the query-weighted nominal-set cumulative regret by
\[
\widetilde{R}_{T}^{(c)}
=
\sum_{s=1}^{m^{(c)}}N_s
\bigl(f(x_s^\dagger)-f(x_s)\bigr).
\]
Then, on \(\mathcal E_f\),
\[
\widetilde{R}_{T}^{(c)}
=
\mathcal{O}\left(
\frac{
(\log T)^{\frac{3(d+1)}{2}}
\log\bigl((\log T)^{d+1}\bigr)
}
{
c\log(1+c^{-2})
}
\right)
+
\mathcal{O}\left(
\frac{1}{c}
\bigl(\log((\log T)^{d+1})\bigr)^2
\right),
\]
for fixed \(c\) and fixed \(\delta\).
\end{theoremone}

\begin{proof}
We work on the objective confidence event \(\mathcal E_f\), which holds with
probability at least \(1-\delta\). On this event, for all \(x\in\mathcal X\)
and all stages \(s\),
\[
f(x)
\le
\mu_{f,s-1}(x)
+
\beta_s^{(f,c)}\sigma_{f,s-1}(x)
=
\mathrm{UCB}_{f,s}(x),
\]
and
\[
f(x)
\ge
\mu_{f,s-1}(x)
-
\beta_s^{(f,c)}\sigma_{f,s-1}(x)
=
\mathrm{LCB}_{f,s}(x).
\]
Assume
\[
\widehat S_s\neq\emptyset,
\]
we select $x_s$ from the nominal safe set
\[
x_s\in\widehat S_s.
\]
Define the simple regret 
\[
\widetilde r_s
=
f(x_s^\dagger)-f(x_s).
\]
Using the objective confidence bounds, we have
\[
f(x_s^\dagger)
\le
\mathrm{UCB}_{f,s}(x_s^\dagger),
\]
and
\[
f(x_s)
\ge
\mathrm{LCB}_{f,s}(x_s).
\]
Therefore,
\[
\widetilde r_s
\le
\mathrm{UCB}_{f,s}(x_s^\dagger)
-
\mathrm{LCB}_{f,s}(x_s).
\]
Adding and subtracting \(\mathrm{UCB}_{f,s}(x_s)\), we obtain
\begin{align}
\widetilde r_s
&\le
\Bigl(
\mathrm{UCB}_{f,s}(x_s^\dagger)
-
\mathrm{UCB}_{f,s}(x_s)
\Bigr)
+
\Bigl(
\mathrm{UCB}_{f,s}(x_s)
-
\mathrm{LCB}_{f,s}(x_s)
\Bigr).
\label{eq:nominal_regret_split}
\end{align}
Since \(x_s\) maximizes \(A_s\) over \(\widehat S_s\), and
\(x_s^\dagger\in\widehat S_s\), we have
\[
A_s(x_s)\ge A_s(x_s^\dagger).
\]
Using
\[
A_s(x)
=
(1-\eta_s)\mathrm{UCB}_{f,s}(x)
-
\eta_s z_s(x),
\]
this implies
\[
(1-\eta_s)\mathrm{UCB}_{f,s}(x_s)
-
\eta_s z_s(x_s)
\ge
(1-\eta_s)\mathrm{UCB}_{f,s}(x_s^\dagger)
-
\eta_s z_s(x_s^\dagger).
\]
Because \(s\ge1\) and \(\eta_s=1/(s+1)\), we have \(1-\eta_s>0\). Rearranging gives
\[
\mathrm{UCB}_{f,s}(x_s^\dagger)
-
\mathrm{UCB}_{f,s}(x_s)
\le
\frac{\eta_s}{1-\eta_s}
\left(
z_s(x_s^\dagger)-z_s(x_s)
\right).
\]
By assumption,
\[
|z_s(x)|\le M_z,
\qquad
\forall x\in\widehat S_s,\ \forall s.
\]
Thus,
\[
z_s(x_s^\dagger)-z_s(x_s)
\le
2M_z.
\]
Also,
\[
\mathrm{UCB}_{f,s}(x_s)
-
\mathrm{LCB}_{f,s}(x_s)
=
2\beta_s^{(f,c)}\sigma_{f,s-1}(x_s).
\]
Substituting these two bounds into \eqref{eq:nominal_regret_split}, we obtain
\[
\widetilde r_s
\le
2\beta_s^{(f,c)}\sigma_{f,s-1}(x_s)
+
\frac{2M_z\eta_s}{1-\eta_s}.
\]
Multiplying simple regret $\widetilde r_s$ by \(N_s\) and summing over \(s=1,\ldots,m^{(c)}\), we obtain
\[
\widetilde R_T^{(c)}
\le
\sum_{s=1}^{m^{(c)}}N_s
\left[
2\beta_s^{(f,c)}\sigma_{f,s-1}(x_s)
+
\frac{2M_z\eta_s}{1-\eta_s}
\right].
\]
We now bound \(\widetilde R_T^{(c)}\). By the precision rule,
\[
\epsilon_s^{(c)}
=
c\frac{\sigma_{f,s-1}(x_s)}{\sqrt{\lambda}},
\]
so
\[
\sigma_{f,s-1}(x_s)
=
\frac{\sqrt{\lambda}}{c}\epsilon_s^{(c)}.
\]
Using the query bound,
\[
N_s
\le
\frac{C_1}{\epsilon_s^{(c)}}
\log\left(
\frac{2\bar m^{(c)}}{\delta}
\right),
\]
we get
\[
\begin{aligned}
\widetilde R_T^{(c)}
&\le
\frac{2C_1\sqrt{\lambda}}{c}
\log\left(
\frac{2\bar m^{(c)}}{\delta}
\right)
\sum_{s=1}^{m^{(c)}}\beta_s^{(f,c)}
\\
&\quad+
2C_1M_z
\log\left(
\frac{2\bar m^{(c)}}{\delta}
\right)
\sum_{s=1}^{m^{(c)}}
\frac{\eta_s}{(1-\eta_s)\epsilon_s^{(c)}}.
\end{aligned}
\]
Since \(\beta_s^{(f,c)}\) is nondecreasing,
\[
\sum_{s=1}^{m^{(c)}}\beta_s^{(f,c)}
\le
m^{(c)}\beta_{m^{(c)}}^{(f,c)}.
\]
Also, using
\[
\sigma_{f,s-1}(x_s)\ge\sigma_{\min},
\]
we have
\[
\frac{1}{\epsilon_s^{(c)}}
=
\frac{\sqrt{\lambda}}
{c\sigma_{f,s-1}(x_s)}
\le
\frac{\sqrt{\lambda}}
{c\sigma_{\min}}.
\]
Therefore,
\[
\begin{aligned}
\widetilde R_T^{(c)}
&\le
\frac{2C_1\sqrt{\lambda}}{c}
m^{(c)}
\beta_{m^{(c)}}^{(f,c)}
\log\left(
\frac{2\bar m^{(c)}}{\delta}
\right)
\\
&\quad+
\frac{2C_1M_z\sqrt{\lambda}}
{c\sigma_{\min}}
\log\left(
\frac{2\bar m^{(c)}}{\delta}
\right)
\sum_{s=1}^{m^{(c)}}
\frac{\eta_s}{1-\eta_s}.
\end{aligned}
\]
Since
\[
\eta_s=\frac{1}{s+1},
\]
we have
\[
\frac{\eta_s}{1-\eta_s}
=
\frac{1}{s}.
\]
Hence
\[
\sum_{s=1}^{m^{(c)}}\frac{\eta_s}{1-\eta_s}
=
\sum_{s=1}^{m^{(c)}}\frac{1}{s}
=
\mathcal{O}\bigl(\log m^{(c)}\bigr).
\]
Using the SE-kernel stage-count bound,
\[
m^{(c)}
\le
\bar m^{(c)}
=
\mathcal{O}\left(
\frac{(\log T)^{d+1}}
{\log(1+c^{-2})}
\right),
\]
and the confidence-width bound,
\[
\beta_{m^{(c)}}^{(f,c)}
=
\mathcal{O}\left(
(\log T)^{\frac{d+1}{2}}
\right),
\]
we obtain, for fixed \(c\) and fixed \(\delta\),
\[
\widetilde R_T^{(c)}
=
\mathcal{O}\left(
\frac{
(\log T)^{\frac{3(d+1)}{2}}
\log\bigl((\log T)^{d+1}\bigr)
}
{
c\log(1+c^{-2})
}
\right)
+
\mathcal{O}\left(
\frac{1}{c}
\bigl(\log((\log T)^{d+1})\bigr)^2
\right).
\]
This completes the proof.
\end{proof}

\section{Proof of Theorem 2}\label{Appdix:c}
\begin{theoremtwo}[Bounded variance regret of Q-Safe BO]
\label{thm:modified_bounded_variance_regret_appdix}
Assume the conditions of
Theorem 1. The variance of the observations is bounded by $\sigma^{2}$.
Then, with probability at least \(1-\delta\),
\[
\begin{aligned}
\widetilde R_T^{(c)}
&\le
\frac{2C_2\sigma\sqrt{\lambda}}{c}\,
m^{(c)}\beta_{m^{(c)}}^{(f,c)}
\mathcal L_\sigma(T)
\log\frac{2\bar m^{(c)}}{\delta}
\\
&\quad
+
\frac{2C_2\sigma M_z\sqrt{\lambda}}{c\sigma_{\min}}\,
\mathcal L_\sigma(T)
\log\frac{2\bar m^{(c)}}{\delta}
H_{m^{(c)}},
\end{aligned}
\]
where
\[
H_{m^{(c)}}=\sum_{s=1}^{m^{(c)}}\frac{1}{s},
\] and \[
\mathcal L_\sigma(T)
:=
\log_2^{3/2}(8\sigma T)
\log_2\!\bigl(\log_2(8\sigma T)\bigr).
\]
Consequently, ignoring the same \(\log(1/\delta)\) factors as in
Theorem~\ref{thm:objective_confidence_regret}, for the SE kernel,
\[
\begin{aligned}
\widetilde R_T^{(c)}
&=
\mathcal O\!\left(
\frac{
\sigma\,(\log T)^{3(d+1)/2}
\log\!\bigl((\log T)^{d+1}\bigr)
\log_2^{3/2}(\sigma T)
\log_2\!\bigl(\log_2(\sigma T)\bigr)
}{
c\,\log(1+c^{-2})
}
\right)
\\
&\quad
+
\mathcal O\!\left(
\frac{
\sigma
\log_2^{3/2}(\sigma T)
\log_2\!\bigl(\log_2(\sigma T)\bigr)
}{
c
}
\bigl(\log\!\bigl((\log T)^{d+1}\bigr)\bigr)^2
\right).
\end{aligned}
\]
For the linear kernel,
\[
\begin{aligned}
\widetilde R_T^{(c)}
&=
\mathcal O\!\left(
\frac{
\sigma\,d^{3/2}(\log T)^{3/2}
\log(d\log T)
\log_2^{3/2}(\sigma T)
\log_2\!\bigl(\log_2(\sigma T)\bigr)
}{
c\,\log(1+c^{-2})
}
\right)
\\
&\quad
+
\mathcal O\!\left(
\frac{
\sigma
\log_2^{3/2}(\sigma T)
\log_2\!\bigl(\log_2(\sigma T)\bigr)
}{
c
}
\bigl(\log(d\log T)\bigr)^2
\right).
\end{aligned}
\]
\end{theoremtwo}

\begin{proof}
We work on the same high-probability event as in
Theorem~\ref{thm:objective_confidence_regret}, on which the
objective confidence bounds hold for all \(x\in\mathcal X\) and all completed
stages \(s\). Hence
\[
f(x)
\le
\mu_{f,s-1}(x)
+
\beta_s^{(f,c)}\sigma_{f,s-1}(x)
=
\mathrm{UCB}_{f,s}(x),
\]
and
\[
f(x)
\ge
\mu_{f,s-1}(x)
-
\beta_s^{(f,c)}\sigma_{f,s-1}(x)
=
\mathrm{LCB}_{f,s}(x).
\]
Define the safe-set simple regret at stage \(s\) by
\[
\widetilde r_s
=
f(x_s^\dagger)-f(x_s).
\]
Using the confidence bounds,
\[
f(x_s^\dagger)
\le
\mathrm{UCB}_{f,s}(x_s^\dagger),
\qquad
f(x_s)
\ge
\mathrm{LCB}_{f,s}(x_s).
\]
Therefore,
\[
\widetilde r_s
\le
\mathrm{UCB}_{f,s}(x_s^\dagger)
-
\mathrm{LCB}_{f,s}(x_s).
\]
Adding and subtracting \(\mathrm{UCB}_{f,s}(x_s)\), we get
\[
\begin{aligned}
\widetilde r_s
&\le
\Bigl(
\mathrm{UCB}_{f,s}(x_s^\dagger)
-
\mathrm{UCB}_{f,s}(x_s)
\Bigr)
\\
&\quad
+
\Bigl(
\mathrm{UCB}_{f,s}(x_s)
-
\mathrm{LCB}_{f,s}(x_s)
\Bigr).
\end{aligned}
\]
Since \(x_s\) maximizes \(A_s\) over \(\widehat S_s\), and since
\(x_s^\dagger\in\widehat S_s\), we have
\[
A_s(x_s)\ge A_s(x_s^\dagger).
\]
Using
\[
A_s(x)
=
(1-\eta_s)\mathrm{UCB}_{f,s}(x)-\eta_s z_s(x),
\]
this gives
\[
(1-\eta_s)\mathrm{UCB}_{f,s}(x_s)
-
\eta_s z_s(x_s)
\ge
(1-\eta_s)\mathrm{UCB}_{f,s}(x_s^\dagger)
-
\eta_s z_s(x_s^\dagger).
\]
Because \(\eta_s=1/(s+1)\), we have \(1-\eta_s>0\). Rearranging,
\[
\mathrm{UCB}_{f,s}(x_s^\dagger)
-
\mathrm{UCB}_{f,s}(x_s)
\le
\frac{\eta_s}{1-\eta_s}
\bigl(
z_s(x_s^\dagger)-z_s(x_s)
\bigr).
\]
Since \(|z_s(x)|\le M_z\) on \(\widehat S_s\),
\[
z_s(x_s^\dagger)-z_s(x_s)
\le
2M_z.
\]
Also,
\[
\mathrm{UCB}_{f,s}(x_s)
-
\mathrm{LCB}_{f,s}(x_s)
=
2\beta_s^{(f,c)}\sigma_{f,s-1}(x_s).
\]
Thus,
\[
\widetilde r_s
\le
2\beta_s^{(f,c)}\sigma_{f,s-1}(x_s)
+
\frac{2M_z\eta_s}{1-\eta_s}.
\]
Since \(\eta_s=1/(s+1)\),
\[
\frac{\eta_s}{1-\eta_s}
=
\frac{1}{s}.
\]
Hence
\[
\widetilde r_s
\le
2\beta_s^{(f,c)}\sigma_{f,s-1}(x_s)
+
\frac{2M_z}{s}.
\]
Multiplying by \(N_s\) and summing over completed stages,
\[
\widetilde R_T^{(c)}
\le
\sum_{s=1}^{m^{(c)}}N_s
\left[
2\beta_s^{(f,c)}\sigma_{f,s-1}(x_s)
+
\frac{2M_z}{s}
\right].
\]
Define
\[
\mathcal L_\sigma(T)
:=
\log_2^{3/2}(8\sigma T)
\log_2\!\bigl(\log_2(8\sigma T)\bigr).
\]
By the bounded variance QMC estimation query bound from
Lemma~\ref{Lemma:QMC}, for every completed stage,
\[
N_s
\le
\frac{C_2\sigma}{\epsilon_s^{(c)}}\,
\mathcal L_\sigma(T)
\log\frac{2\bar m^{(c)}}{\delta}.
\]
Therefore,
\[
\begin{aligned}
\widetilde R_T^{(c)}
&\le
2C_2\sigma\mathcal L_\sigma(T)
\log\frac{2\bar m^{(c)}}{\delta}
\sum_{s=1}^{m^{(c)}}
\beta_s^{(f,c)}
\frac{\sigma_{f,s-1}(x_s)}{\epsilon_s^{(c)}}
\\
&\quad
+
2C_2\sigma M_z\mathcal L_\sigma(T)
\log\frac{2\bar m^{(c)}}{\delta}
\sum_{s=1}^{m^{(c)}}
\frac{1}{s\,\epsilon_s^{(c)}}.
\end{aligned}
\]
Using the precision rule
\[
\epsilon_s^{(c)}
=
c\frac{\sigma_{f,s-1}(x_s)}{\sqrt{\lambda}},
\]
we have
\[
\frac{\sigma_{f,s-1}(x_s)}{\epsilon_s^{(c)}}
=
\frac{\sqrt{\lambda}}{c}.
\]
Thus the first sum becomes
\[
2C_2\sigma\mathcal L_\sigma(T)
\log\frac{2\bar m^{(c)}}{\delta}
\sum_{s=1}^{m^{(c)}}
\beta_s^{(f,c)}
\frac{\sigma_{f,s-1}(x_s)}{\epsilon_s^{(c)}}
=
\frac{2C_2\sigma\sqrt{\lambda}}{c}
\mathcal L_\sigma(T)
\log\frac{2\bar m^{(c)}}{\delta}
\sum_{s=1}^{m^{(c)}}\beta_s^{(f,c)}.
\]
Since \(\beta_s^{(f,c)}\) is nondecreasing,
\[
\sum_{s=1}^{m^{(c)}}\beta_s^{(f,c)}
\le
m^{(c)}\beta_{m^{(c)}}^{(f,c)}.
\]
Therefore,
\[
\frac{2C_2\sigma\sqrt{\lambda}}{c}
\mathcal L_\sigma(T)
\log\frac{2\bar m^{(c)}}{\delta}
\sum_{s=1}^{m^{(c)}}\beta_s^{(f,c)}
\le
\frac{2C_2\sigma\sqrt{\lambda}}{c}
m^{(c)}\beta_{m^{(c)}}^{(f,c)}
\mathcal L_\sigma(T)
\log\frac{2\bar m^{(c)}}{\delta}.
\]
For the second sum, again using
\[
\epsilon_s^{(c)}
=
c\frac{\sigma_{f,s-1}(x_s)}{\sqrt{\lambda}},
\]
we have
\[
\frac{1}{\epsilon_s^{(c)}}
=
\frac{\sqrt{\lambda}}{c\,\sigma_{f,s-1}(x_s)}.
\]
By the assumption \(\sigma_{f,s-1}(x_s)\ge\sigma_{\min}\),
\[
\frac{1}{\epsilon_s^{(c)}}
\le
\frac{\sqrt{\lambda}}{c\sigma_{\min}}.
\]
Hence
\[
\sum_{s=1}^{m^{(c)}}\frac{1}{s\,\epsilon_s^{(c)}}
\le
\frac{\sqrt{\lambda}}{c\sigma_{\min}}
\sum_{s=1}^{m^{(c)}}\frac{1}{s}
=
\frac{\sqrt{\lambda}}{c\sigma_{\min}}H_{m^{(c)}}.
\]
Substituting this bound gives
\[
2C_2\sigma M_z\mathcal L_\sigma(T)
\log\frac{2\bar m^{(c)}}{\delta}
\sum_{s=1}^{m^{(c)}}
\frac{1}{s\,\epsilon_s^{(c)}}
\le
\frac{2C_2\sigma M_z\sqrt{\lambda}}{c\sigma_{\min}}
\mathcal L_\sigma(T)
\log\frac{2\bar m^{(c)}}{\delta}
H_{m^{(c)}}.
\]
Combining the two estimates, we obtain
\[
\begin{aligned}
\widetilde R_T^{(c)}
&\le
\frac{2C_2\sigma\sqrt{\lambda}}{c}
m^{(c)}\beta_{m^{(c)}}^{(f,c)}
\mathcal L_\sigma(T)
\log\frac{2\bar m^{(c)}}{\delta}
\\
&\quad
+
\frac{2C_2\sigma M_z\sqrt{\lambda}}{c\sigma_{\min}}
\mathcal L_\sigma(T)
\log\frac{2\bar m^{(c)}}{\delta}
H_{m^{(c)}}.
\end{aligned}
\]
It remains to substitute the kernel-specific bounds. For the SE kernel,
\[
m^{(c)}
\le
\bar m^{(c)}
=
\mathcal O\!\left(
\frac{(\log T)^{d+1}}{\log(1+c^{-2})}
\right),
\]
and
\[
\beta_{m^{(c)}}^{(f,c)}
=
\mathcal O\!\left(
(\log T)^{(d+1)/2}
\right).
\]
Moreover,
\[
H_{m^{(c)}}=\mathcal O(\log m^{(c)})
\]
and, ignoring \(\log(1/\delta)\) factors,
\[
\log\frac{2\bar m^{(c)}}{\delta}
=
\mathcal O\!\left(
\log\!\bigl((\log T)^{d+1}\bigr)
\right).
\]
Therefore,
\[
\begin{aligned}
\widetilde R_T^{(c)}
&=
\mathcal O\!\left(
\frac{
\sigma\,(\log T)^{3(d+1)/2}
\log\!\bigl((\log T)^{d+1}\bigr)
\log_2^{3/2}(\sigma T)
\log_2\!\bigl(\log_2(\sigma T)\bigr)
}{
c\,\log(1+c^{-2})
}
\right)
\\
&\quad
+
\mathcal O\!\left(
\frac{
\sigma
\log_2^{3/2}(\sigma T)
\log_2\!\bigl(\log_2(\sigma T)\bigr)
}{
c
}
\bigl(\log\!\bigl((\log T)^{d+1}\bigr)\bigr)^2
\right).
\end{aligned}
\]
For the linear kernel,
\[
m^{(c)}
\le
\bar m^{(c)}
=
\mathcal O\!\left(
\frac{d\log T}{\log(1+c^{-2})}
\right),
\]
and
\[
\beta_{m^{(c)}}^{(f,c)}
=
\mathcal O\!\left(
d^{1/2}(\log T)^{1/2}
\right).
\]
Thus,
\[
\begin{aligned}
\widetilde R_T^{(c)}
&=
\mathcal O\!\left(
\frac{
\sigma\,d^{3/2}(\log T)^{3/2}
\log(d\log T)
\log_2^{3/2}(\sigma T)
\log_2\!\bigl(\log_2(\sigma T)\bigr)
}{
c\,\log(1+c^{-2})
}
\right)
\\
&\quad
+
\mathcal O\!\left(
\frac{
\sigma
\log_2^{3/2}(\sigma T)
\log_2\!\bigl(\log_2(\sigma T)\bigr)
}{
c
}
\bigl(\log(d\log T)\bigr)^2
\right).
\end{aligned}
\]
This completes the proof.
\end{proof}

\section{Additional Simulation Study Results}
\label{Appdix:singlerun}
\begin{figure}[H]
    \centering
    \includegraphics[width=\columnwidth]{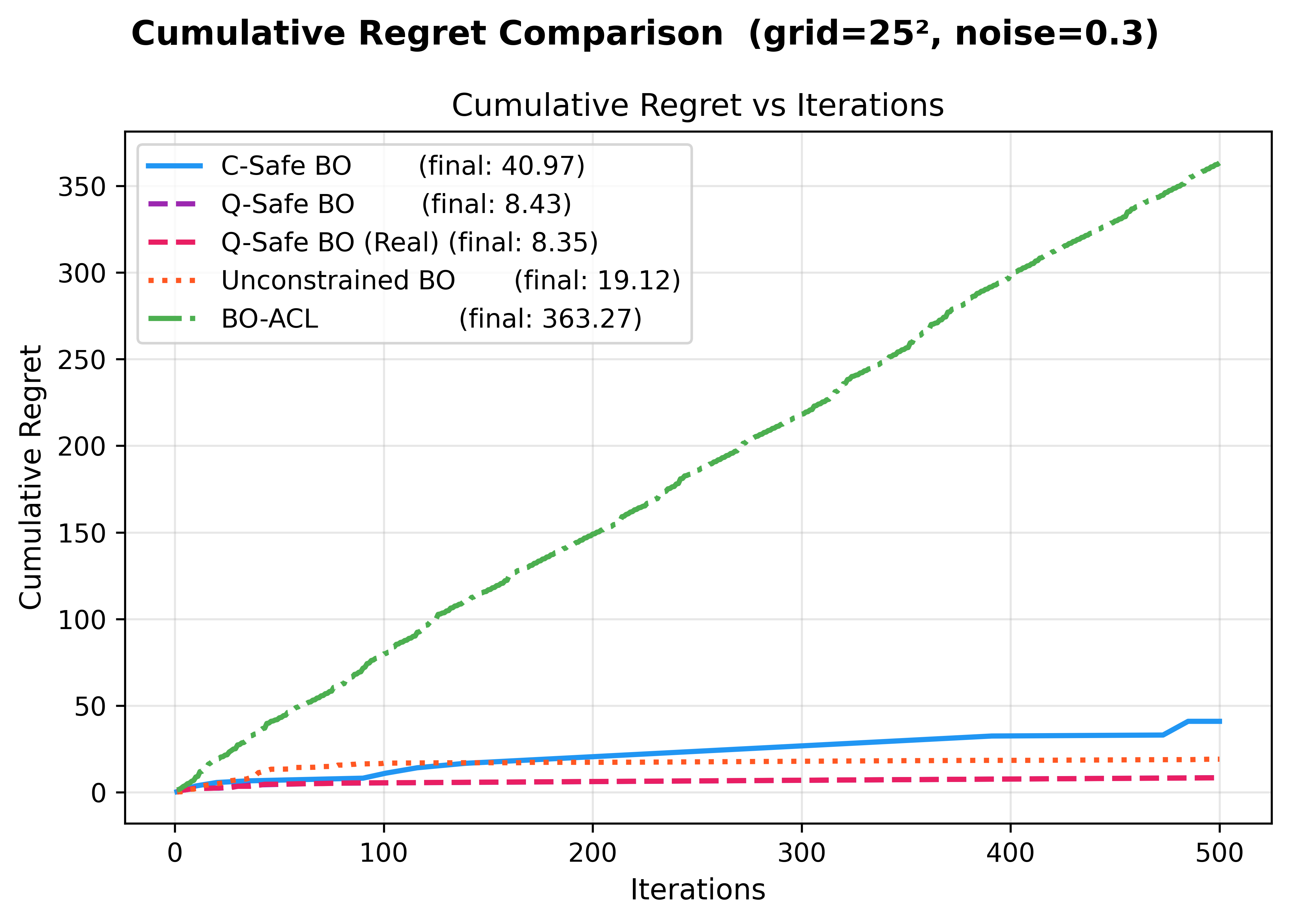}
    \caption{Cumulative regret under the single-run setting.}
    \label{fig:single_seed_maps_regret}
\end{figure}

We show one experimental trial with a \(25\times25\) grid, 500 iterations, and an objective noise standard deviation of 0.3. The numerical results are summarized in Tab. ~\ref{tab:single_seed_simulated}. In addition, Fig. ~\ref{fig:single_seed_maps_regret} reports the cumulative regret curves, while Fig. ~\ref{fig:single_seed_maps} visualizes the query locations and learned feasibility regions.

\begin{table}[H]
\centering
\scriptsize
\setlength{\tabcolsep}{4pt}
\renewcommand{\arraystretch}{1.05}

\begin{adjustbox}{max width=\columnwidth}
\begin{tabular}{
l
S[table-format=-1.3]
S[table-format=1.3]
S[table-format=3.3]
S[table-format=1.3]
S[table-format=1.3]
}
\toprule
Method 
& {\makecell{Best\\Safe}} 
& {\makecell{Simple\\Regret}} 
& {\makecell{Cumul.\\Regret}} 
& {\makecell{Safe\\Rate}} 
& {\makecell{Viol.\\Rate}} \\
\midrule
C-Safe BO              & {\bfseries -0.979} & {\bfseries 0.000} & 40.973  & 1.000 & 0.000 \\
Q-Safe BO            & {\bfseries -0.979} & {\bfseries 0.000} & 8.430 & 1.000 & 0.000 \\
Q-Safe BO (Real)     & {\bfseries -0.979} & {\bfseries 0.000} & {\bfseries 8.353}  & 1.000 & 0.000 \\
Unconstr. BO         & -0.779 & 0.200 & 19.117  & 0.000 & 1.000 \\
BO-ACL               & -0.842 & 0.137 & 363.270 & 0.736 & 0.264 \\
\bottomrule
\end{tabular}
\end{adjustbox}
\caption{Single trial performance comparison on the two-dimensional constrained synthetic benchmark under a fixed computational budget of 500 optimization iterations. The global optimum is $-0.979$.}
\label{tab:single_seed_simulated}
\end{table}

Tab.~\ref{tab:single_seed_simulated} shows that Q-Safe BO reaches the feasible
global optimum while maintaining a 100\% safe rate. Specifically, Q-Safe BO (Real) achieves the lowest cumulative regret and zero simple regret at the end stage among the safe methods. Compared with C-Safe BO, Q-Safe BO reduces cumulative regret from 40.973 to 8.43 (simulator) and 8.353 (real quantum hardware). This result indicates that QMC estimation improves
the sample efficiency under the same number of
optimization iterations. 

Fig. \ref{fig:single_seed_maps_regret} further illustrates the cumulative regret behavior over the full optimization process. C-Safe BO also preserves feasibility, but its regret increases rapidly during the middle of the optimization process. Both Q-Safe BO and Q-Safe BO (Real) remain competitive and achieve lower cumulative regret than C-Safe BO. In contrast, BO-ACL accumulates regret much more rapidly, indicating that its constraint-learning strategy is less effective for this benchmark.
Although unconstrained BO has relatively low cumulative regret, this result should be treated with caution. Since unconstrained BO does not model the safety requirement, it queries infeasible regions throughout the optimization process. This is reflected by its violation rate of 1.000 in Tab. ~\ref{tab:single_seed_simulated}. Therefore, its low cumulative regret does not correspond to a feasible optimization trajectory.

% ------------------------------------------------------------
% c-scaled variant of Q-GP-UCB
% ------------------------------------------------------------

% \if0\blind{
% \section*{Acknowledgements}
% The authors acknowledge the generous support from the IBM-RPI FCRC Research Grant on this project.} \fi

\if0\blind{
\section*{Data Availability Statement}
The data supporting the findings of this study will be made available from the corresponding author upon reasonable request after the publication of this paper.} \fi

\bibliographystyle{chicago}
\spacingset{1}
\bibliography{IISE-Trans}
	
\end{document}